\documentclass[english,11pt]{elsarticle} 
\usepackage[T1]{fontenc}
\usepackage{float}
\usepackage{amsmath}
\usepackage{amsthm}
\usepackage{amssymb}
\usepackage{amsfonts}
\usepackage{bbm}
\allowdisplaybreaks
\usepackage{graphicx}
\usepackage{natbib}
\setcitestyle{square, comma, numbers,sort&compress, super}
\usepackage{float}
\usepackage{subfloat}
\usepackage{epstopdf}
\usepackage{hyperref}
\hypersetup{colorlinks,allcolors=black}
\usepackage{nomencl}
\makenomenclature
\usepackage{algorithm}
\usepackage{algorithmic}

\usepackage{url}
\usepackage{graphicx}
\graphicspath{{images/}} 
\usepackage{comment}
\usepackage{ragged2e}
\usepackage{pgfplots}
\pgfplotsset{compat=newest}
\usetikzlibrary{plotmarks}
\usetikzlibrary{arrows.meta}
\usepgfplotslibrary{patchplots}
\usepackage{grffile}
\usepackage{amsmath}
\usepackage{autobreak}
\usepackage{appendix} 
\usepackage{appendix}
\usepackage[symbol]{footmisc}

\usepackage{paralist}
\usepackage{tablefootnote}
\usepackage{booktabs}
\usepackage{nomencl}
\usepackage{amsmath}
\usepackage{array}
\makenomenclature

\usepackage{epsfig}
\usepackage{subfigure}
\usepackage{amssymb,amsmath,amsfonts,layout,graphicx}
\usepackage{makeidx}
\usepackage{babel}
\usepackage{tikz}
\usepackage{sublabel}
\usepackage{cases}
\usepackage{algorithm}
\usepackage{algorithmic}

\usepackage
[
        letterpaper,
        left=1.91cm,
        right=1.91cm,
        top=1.91cm,
        bottom=2cm,
]
{geometry}

\newtheorem{remark}{Remark}

\DeclareMathOperator*{\argmax}{arg\,max}

\biboptions{numbers,sort&compress} 

\renewcommand{\nomname}{Notations}
\renewcommand{\nomgroup}[1]{%
  \item[\bfseries
  \ifstrequal{#1}{S}{Set}{%
  \ifstrequal{#1}{F}{Function}{%
  \ifstrequal{#1}{P}{Parameter}{%
  \ifstrequal{#1}{D}{Decision Variable}{}}}}]
}

\begin{document}
\begin{frontmatter}
\title{\textbf{Coordinating Ride-Pooling with Public Transit using Reward-Guided Conservative Q-Learning: An Offline Training and Online \\Fine-Tuning Reinforcement Learning Framework}}	

\author[1staddress]{Yulong Hu}
\ead{yhucm@connect.ust.hk}
\author[1staddress]{Tingting Dong}
\ead{dongtthit@ust.hk}
\author[1staddress,2ndaddress]{Sen Li\corref{mycorrespondingauthor}}
\cortext[mycorrespondingauthor]{Corresponding author}
\ead{cesli@ust.hk}
\address[1staddress]{Department of Civil and Environmental Engineering, The Hong Kong University of Science and Technology, Hong Kong, China}
\address[2ndaddress]{Intelligent Transportation Thrust, Systems Hub, The Hong Kong University of Science and Technology (Guangzhou), Guangzhou, China}

\begin{abstract}
This paper introduces a novel reinforcement learning (RL) framework, termed Reward-Guided Conservative Q-learning (RG-CQL), to enhance coordination between ride-pooling and public transit within a multimodal transportation network. We model each ride-pooling vehicle as an agent governed by a Markov Decision Process (MDP), which includes a state for each agent encompassing the vehicle's location, the number of vacant seats, and all pertinent information regarding the passengers on board. We propose an offline training and online fine-tuning RL framework to learn the optimal operational decisions of the multimodal transportation systems, including rider-vehicle matching, selection of drop-off locations for passengers, and vehicle routing decisions, with improved data efficiency. During the offline training phase, we develop a Conservative Double Deep Q Network (CDDQN) as the action executor and a supervised learning-based reward estimator, termed the Guider Network, to extract valuable insights into action-reward relationships from data batches. In the online fine-tuning phase, the Guider Network serves as an exploration guide, aiding CDDQN in effectively and conservatively exploring unknown state-action pairs to bridge the gap between the conservative offline training and optimistic online fine-tuning. The efficacy of our algorithm is demonstrated through a realistic case study using real-world data from Manhattan. We show that integrating ride-pooling with public transit outperforms two benchmark cases—solo rides coordinated with transit and ride-pooling without transit coordination—by 17\% and 22\% in the achieved system rewards, respectively. Furthermore, our innovative offline training and online fine-tuning framework offers a remarkable 81.3\% improvement in data efficiency compared to traditional online RL methods with adequate exploration budgets, with a 4.3\% increase in total rewards and a 5.6\% reduction in overestimation errors. {Experimental results further demonstrate that RG-CQL effectively addresses the challenges of transitioning from offline to online RL in large-scale ride-pooling systems integrated with transit.}
\end{abstract}

\begin{keyword}
Multimodal Transportation, Ride-Pooling, Public Transit, Offline Reinforcement Learning, Online Fine-tuning, Safe Exploration, Reward Model.
\end{keyword}
\end{frontmatter}

\section{Introduction}

In recent years, the widespread use of transportation network companies, such as Uber, Lyft, and Didi, has raised concerns about their negative impacts on traffic and the environment. For these ride-hailing platforms, the ability to immediately respond to passenger requests is a crucial element of their success. However, this business model heavily relies on a large number of drivers who remain idle and cruise around waiting for passengers \cite{schaller2017empty}. This not only leads to low vehicle utilization rates and high fuel costs but also exacerbates traffic congestion and carbon emission in urban areas. Furthermore, a more pressing concern is that ride-hailing services may compete with public transit systems. The convenience offered by ride-hailing services can divert a substantial number of riders away from public transportation~\cite{chen2018pricing, wang2016approximating, wang2019routing,qin2022government}, leading to what could become a vicious cycle that undermines public transit development. This may contribute to a rise in vehicle miles traveled, increased traffic congestion, reduced transit ridership, and higher emissions, which run counter to the philosophy of developing transit-oriented, multimodal transportation systems that are operationally efficient, financially sustainable, and environmentally friendly.

To address the aforementioned concerns, substantial research has been conducted along two major directions, including: (a) how to improve the operational efficiency of ride-hailing services; and (b) how to promote collaborative relationships between ride-hailing and public transportation systems. To achieve the first goal, a significant body of work has focused on ride-pooling services, which allow multiple passengers to share a ride in a single vehicle through a digital platform, traveling to similar or conveniently aligned destinations. By consolidating multiple orders within the same vehicle, ride-pooling can effectively enhance vehicle utilization rates, while also reducing per-person travel miles and carbon emissions. In this vein, various forms of ride-pooling services have been explored, including multi-hop ride-pooling \cite{singh2021distributed}, diurnal-adaptive ride-pooling management \cite{haliem2021adapool}, and ride-pooling with passenger transfers \cite{wang2023optimization}. To achieve the second objective, another stream of research has investigated the coordination between ride-hailing services and public transit systems~\cite{feng2022coordinating,gao2024regulating,xu2024design,wang2024coordinative}. The underlying premise is that ride-hailing services can either complement or compete with public transit, depending on their operational approach. To maximize the complementary effects between these two, strategies have been proposed to utilize ride-hailing services for the first and/or last leg of a journey, while leveraging public transit for the long-distance middle leg between transport hubs. This approach could potentially increase transit ridership by extending public transit access to those who are unable to reach transit facilities without the aid of ride-hailing services.

There exists significant synergy by combining the aforementioned two research directions, where ride-pooling services operate in conjunction with transit networks, enabling passengers to share a ride on the ride-hailing platform to cover the first and last miles of their trip, while public transportation handles the long-distance segments between transportation hubs. This arrangement is mutually beneficial for both ride-pooling and public transportation networks. In particular, a crucial determinant of the performance for ride-pooling services is the shareability of rides, which depends on whether the origins and destinations of passengers are compatible and can thus be pooled together. By integrating with transit services, we can allow significant flexibility for the ride-pooling platform to choose the drop-off point, which can significantly enhance the chances of passengers being pooled. Meanwhile, pooled rides can reduce costs for passengers and enable more people, especially lower-income passengers who are more transit-dependent, to access transit networks. This can initiate a virtuous circle for transit development, where increased ridership improves the financial condition of transit operators, who can then further enhance service quality and in return attract more passengers.


However, operating ride-pooling services within multimodal transportation networks that consider inter-modal transfers poses significant research challenges. Firstly, the decision-making involved in managing large-scale multimodal networks under uncertainties is complex for ride-hailing platforms. Specifically, the platform must determine how to pool passengers, match them to vehicles, route these vehicles, and select appropriate pick-up and drop-off points for inter-modal transfers. These tasks are intricately intertwined and must be addressed in real-time amid significant uncertainties, rendering traditional model-based algorithms \cite{stiglic2018enhancing, ma2019dynamic, gu2024algorithms} inadequate. This has led to the exploration of Reinforcement Learning (RL) algorithms. However, in the realm of RL, especially when applied to the coordinated management of ride-pooling and transit networks, training models from scratch is not only time-consuming but also prone to yielding low-quality local solutions due to the enormous decision space and the trial-and-error, data-intensive nature of RL algorithms \cite{sutton1998introduction, kumar2020conservative} within the multimodal transportation system. In this context, agents must continuously interact with the environment or simulator throughout the training process, which can be exceedingly time-consuming, and this challenge is further magnified in scenarios involving thousands of agents with sophisticated operational strategies. To address these issues, some transportation researchers have proposed adopting off-policy RL techniques \cite{yu2022batch, tang2019deep, jiao2021real, wei2023reinforcement}, such as TD learning, to learn value functions from batches of past vehicle transitions. However, these methods often suffer from significant overestimation issues due to out-of-distribution (OOD) data transitions during the offline training phase \cite{levine2020offline}. Moreover, there is a persistent gap between offline and online learning, and current offline RL methods also experience slow learning rates during subsequent online training \cite{nakamoto2024cal}, further limiting the practical application of these approaches.


Acknowledging the challenges mentioned above and inspired by the recent progress of reward models in enabling Large Language Models (LLMs) to learn from human feedback through RL \cite{ziegler2019fine, ouyang2022training, rafailov2024direct} {and rewarding mechanism in ride-sharing to promote acceptable rides \cite{wang2018stable,agatz2011dynamic,hsieh2023improving,hsieh2024comparison}}, we hereby propose Reward Guided Conservative Q-learning (\textbf{RG-CQL}). This approach leverages an offline training and online fine-tuning RL framework to effectively coordinate ride-pooling with transit, complemented with a Guider to improve the efficiency of exploration. In particular, we first formulate the ride-pooling problem in multimodal transportation networks as a Markov Decision Process (MDP), and then proceed to train an offline learning Conservative Double Deep Q Network (CDDQN) \cite{van2016deep, kumar2020conservative} policy, which serves as the potential action executor, alongside a supervised learning reward estimator  during the offline training phase to maximize insights from batches of past environmental transitions. Subsequently, during the online fine-tuning phase, we employ the reward estimator as the exploration guide, aiding the CDDQN in effectively and conservatively exploring unknown state-action pairs. Our extensive numerical experiments validate that the proposed RG-CQL framework markedly enhances the operational performance of multimodal transportation systems. The major contributions of our paper can be summarized as follows:

\begin{itemize}
    \item We addressed the coordinated management of ride-pooling with public transit under a RL framework, which is absent in the existing literature. Specifically, we modeled each ride-pooling vehicle as an agent operating under an MDP. The state of each agent includes the vehicle's location, the number of vacant seats, and all pertinent information regarding the passengers currently on board. Actions within this framework determine the drop-off locations—either transit stations or final destinations—for each onboard passenger. In this model, agents independently decide the drop-off locations for each matched passenger, while the rider-bundling and rider-vehicle matching is addressed centrally through bipartite matching, with edges weighted according to value functions derived from the MDP. With decisions about matching and drop-offs resolved, a separate vehicle routing problem can be independently solved and dynamically adjusted in real-time for each ride-pooling vehicle. Overall, our MDP formulation allows us to jointly determine the optimal matches between riders and vehicles, the optimal drop-off locations for each passenger, and the optimal routes for each vehicle, all under  uncertainties within the multimodal transportation system.

    \item We propose a pioneering Reward Guided Conservative Q-learning (RG-CQL) framework that integrates offline training with online fine-tuning to jointly optimize the operation of ride-pooling and public transit services. In the offline training, we employ CDDQN trained via offline RL phase based on historical environmental transitions, complemented by training a Guider that utilizes reward regression. Subsequently, in the online fine-tuning phase, the CDDQN acts as an action executor, while the Guider serves as an exploration guide, helping to filter out irrelevant actions and thereby enhancing the efficiency of the exploration process. This computational framework significantly enhances the system's ability to explore unfamiliar state-action pairs both effectively and cautiously. It effectively bridges the gap between conservative offline training and optimistic online fine-tuning, mitigating issues like initial unlearning and slow adaptation rates. This not only drastically improves training sample efficiency but also reduces the risk of getting stuck in low-quality local solutions. 
    
    \item The proposed algorithm is validated in a realistic case study using real-world data from Manhattan's TNC orders, road networks, and metro systems. Under various ride-sourcing and ride-pooling configurations, we demonstrate that the ride-pooling with transit mode, assuming reasonable customer tolerance for additional detours, surpasses both the case of solo rides coordinated with transit and the ride-pooling only mode by 17\% and 22\% in overall performance, respectively. Moreover, our innovative offline training and online fine-tuning framework shows a remarkable 81.3\% improvement in data efficiency compared to traditional online RL pipelines with adequate exploration budgets. It also achieves a 4.3\% improvement in total rewards and a 5.6\% reduction in overestimation errors. Additionally, our framework surpasses baseline method trained from scratch without sufficient guided exploration by 10.7\%, underscoring the effectiveness of the Guider in enhancing RL performance within the context of managing ride-pooling services in conjunction with public transit in multimodal transportation systems. Our RG-CQL also manages to mitigate the pessimism-optimism gap between offline RL and online fine-tuning, over-performing offline RL method and State-of-the-art offline to online RL baselines in terms of overall performance and training efficiency.
\end{itemize}

The remainder of this paper is structured as follows: Section \ref{sec_literature} reviews relevant studies. Section \ref{sec_prosetup} details the problem setup for coordinated ride-pooling and transit services. Model formulations, including an MDP model and an order-matching model, are presented in Section \ref{sec_formulation}. Section \ref{sec_RL} delves into our RG-CQL framework, discussing tasks in both offline and online learning phases and providing a summary of the algorithm. Section \ref{sec_simulation} discusses simulation experiments and results. Finally, conclusions follow in Section \ref{sec_conclusion}.

\section{Literature Review}\label{sec_literature}
\subsection{Order Matching for Ride-Pooling Platforms} 

Ride-pooling services enable individual passengers to share a ride in a single vehicle, coordinated through a digital platform, to travel to similar or conveniently aligned destinations. The operational dynamics of ride-pooling have garnered considerable attention due to their promising yet unpredictable real-time demand, as evidenced by various studies \cite{Ma2013TshareAL,alonso2017demand,simonetto2019real, ke2023supply}. The nature of this uncertainty, coupled with the full potential of ride-pooling systems, introduces complexity into the process of coordinating vehicles with multiple passengers. Effective coordination requires not only addressing the needs of current passengers but also anticipating the needs of future riders, which includes managing new ride requests and those already being served. Initial investigations in this field have considered short-sighted, or myopic, policies that make vehicular assignments based on presently available information \cite{Ma2013TshareAL,alonso2017demand}. To exemplify, \cite{alonso2017demand} notably advances this by introducing the shareability graph, which identifies possible sharing opportunities between new requests and vehicles on standby. They put forward a batch-matching strategy and crafted a sequential method that divides the decision-making process into vehicle routing and passenger assignment tasks. For more efficient real-time operations, \cite{simonetto2019real} reduces the complexity of the matching problem by limiting the process to pairing a single passenger with a vehicle at each time step. Recent advancements in decision-making processes have increasingly focused on integrating uncertainties related to future demand more effectively through methods like Model Predictive Control (MPC) \cite{8206203,tsao2019model,ali2023rebalancing}, Approximate Dynamic Programming (ADP) \cite{shah2020neural,yu2019integrated,you2024approximate}, and Stochastic Programming \cite{luo2023efficient}. Specifically, MPC optimizes control inputs by looking ahead and forecasting future system behavior over a set time horizon; however, its effectiveness heavily relies on the accuracy of model predictions and necessitates substantial computational resources for real-time operations. ADP tackles the challenge by employing complex representation algorithms to approximate value functions, making its performance heavily dependent on the quality of the approximation model used. Stochastic Programming, on the other hand, introduces randomness into optimization models to better manage uncertainties about the future. While this method provides a robust framework, it is hampered by significant computational complexity, especially as the scale of the problem and the number of uncertainty sources grow. {\em Hence, although these methods above theoretically enhance decision support precision, they confront practical hurdles, including the difficulty for real-time computations and requirements for accurate prior knowledge to formulate high-fidelity models and  encode uncertainty.}
    
To fully address the challenges associated with long-term uncertainties inherent in both the ride-pooling mode and order distributions and waive the need for specific modelling, the RL method shows promise due to its ability to learn from uncertainties through experience, its adaptability in managing large-scale and complex decision spaces, and its strong performance in recent applications \cite{mnih2015human,vinyals2019grandmaster,berner2019dota,ouyang2022training}. Inspired by its potential, \cite{al2019deeppool} initially proposed the use of the DDQN RL method for ride-pooling operations, training it from scratch through online iterative interactions with a specially designed simulator. This enabled individual vehicles to independently learn and adapt dispatch and relocation strategies, demonstrating significant improvements in efficiency and performance over traditional methods. Building on this foundation, subsequent research has expanded online RL training to more complex ride-pooling scenarios. Specifically, \cite{singh2021distributed} introduces a distributed, model-free algorithm that utilizes deep reinforcement learning for multi-hop ride-pooling, leading to significant cost reductions and enhanced fleet utilization. \cite{haliem2021adapool} presents an adaptive, model-free deep reinforcement learning framework for optimizing ride-pooling operations, which leverages online Dirichlet change point detection to adapt to dynamic environments. Additionally, \cite{wang2023optimization} explores an optimization strategy for ride-pooling with passenger transfers using a hybrid model that combines deep RL and ILP. {\em Despite these advancements, training RL from scratch remains challenging, inefficient, and prone to local optima in more complex ride-pooling applications, such as coordinating ride-pooling with transit, due to the trial-and-error, data-intensive nature of RL and limited exploration budgets.}

\subsection{Coordinating Ride-hailing Services with Public Transit}
Another concern accompanying the development of ride-hailing studies is its negative externalities on urban transportation systems, where the most significant issue is the competitive interaction between ride-pooling and public transit. Research has shown that ride-sourcing platforms can replace transit rides, leading to increased vehicle miles traveled, exacerbated traffic congestion, reduced transit ridership, and heightened emissions \cite{chen2018pricing, wang2016approximating, wang2019routing,qin2022government}. To harmonize the ride-hailing services with public transit, several researchers attempt to coordinately dispatch ride-sourcing with public transit \cite{feng2022coordinating,gao2024regulating,xu2024design,wang2024coordinative}. Specifically, Feng et al.~\cite{feng2022coordinating} focused on the first-mile problem with ride-sourcing services and designed an Integer Linear Programming (ILP) Framework to coordinate ride-sourcing with metro systems. To take future uncertainty into consideration within ILP, they aim to estimate long-term expected rewards through TD learning with Online RL. \cite{gao2024regulating} developed a game-theoretical framework to compute the Nash Equilibrium within the intimate interactions between the ride-hailing platform, the transit agency, and multi-class passengers with distinct income levels. Based on the proposed model and solution, they investigated two regulatory policies that can improve transport equity. Xu et al.~\cite{xu2024design} explored the strategic placement of transportation hubs in urban areas, proposing that Transportation Network Companies (TNCs) allocate Autonomous Vehicles to these hubs. Their study models five distinct trip types: Intra-Hub, Inter-Hub, Hub to Outer, Outer to Hub, and Outer to Outer. It also incorporates modelling factors such as passenger choice and waiting times. To solve the proposed model, they employed a Multi-Objective Adaptive Particle Swarm Optimization (MOAPSO) approach to search and find the optimal solutions. \cite{wang2024coordinative} introduced subsidies and additional bus services for passengers in high-demand areas, encouraging them to initially use other forms of public transportation to reach less congested areas before completing their journey via ride-sourcing. To determine the optimal dispatching and subsidy schemes, they developed a bi-level mixed integer programming model grounded in network flow theory and designed a corresponding iterative algorithm to effectively solve the model. {\em Although the works above well study the problem of coordinated dispatch of ride-sourcing and public transit, their models and solutions are mainly focused on solo-rides, and thus might not generalize well to the case of ride-pooling within the context of multimodal transportation networks. }

Aside from solo-rides, ride-pooling services has also been considered in the context of multimodal transportation systems. Along this direction, one stream of works considers the strategic planning decisions in the integrated system comprising both ride-pooling and public transit services. For instance, Zhu et al.~\cite{zhu2020analysis} initially viewed ride-pooling services as both a complementary first- and last-mile cooperator and a competitive alternative to traditional public transit. They developed a multi-modal network model to examine commuter behaviors across various transport modes. Their findings revealed that, under conditions of low fare ratios, a significant number of public transit users might switch to ride-pooling for longer distances. They also noted that prioritizing ride-pooling excessively could reduce its demand for long distance ride-pooling and thus increasing usage of public transit. Liu et al.~\cite{liu2021mobility} proposed an integrated system that positions ride-pooling services as a localized demand-responsive transportation component, solely serving as a first- and last-mile feeder for public transit. They developed aspatial queuing model for demand responsive transportation component and integrate it with transit network to demonstrate the promising performance of the proposed system. \cite{fan2024optimal} introduced a ``hold-dispatch" strategy for ride-pooling operations. In this strategy, ride-pooling vehicles are held until they accumulate a target number of ride requests. Once the threshold is met, the vehicles are dispatched to pick up passengers, following routes optimized by solving the Traveling Salesman Problem (TSP). This strategy aims to maximize overall system efficiency and costs, thereby enhancing the overall effectiveness of ride-pooling as an integral component of public transit systems. 

Another research stream concentrates on operational aspect for coordination  of ride-pooling services in conjunction with public transit. This involves dynamically pooling passengers, matching them with available vehicles, intelligently routing these vehicles, and strategically selecting pick-up and drop-off points for seamless inter-modal transfers. This complex array of tasks requires sophisticated, long-term optimal, and real-time decision-making to ensure efficient and effective transportation service integration. To address existing gaps, \cite{ma2019dynamic} investigated the efficacy of online solution algorithms that utilize queueing-theoretic approaches for vehicle dispatch and idle vehicle relocation, tailored specifically to these challenges. Additionally, \cite{stiglic2018enhancing} explored the potential of integrating ride-pooling with public transit, with a particular focus on the corresponding ride-matching technologies. To enhance the practical deployment of these solutions, \cite{gu2024algorithms} introduced an ILP formulation using a hypergraph representation of the problem and employed approximation algorithms to mitigate computational demands. However, none of the aforementioned works have fully addressed the gap in optimal operational strategies for multimodal transportation networks comprising both ride-pooling and public transit services. Specifically, \cite{ma2019dynamic} utilized a queuing-theoretic model that overlooks the intricate real-time matching decisions required to pair riders, who have distinct origins and destinations, with vehicles that may already carry passengers headed to specific destinations. In contrast, both \cite{stiglic2018enhancing} and \cite{gu2024algorithms} considered detailed real-time matching; however, the algorithms proposed in these studies are predominantly myopic, failing to consider the temporal correlations between current and future decisions, which are uncertain and might not be revealed at the time of decision-making.


\subsection{Offline RL and its Deployment in Ride-Sourcing Dispatch}
To mitigate the sample complexity inherent in the trial-and-error and data-intensive nature of online RL, offline RL, or batch RL, aims to train agents directly from pre-collected datasets instead of active interactions with the environment. This approach seeks to align RL with the data and time efficiency characteristic of supervised learning. Observing the promising developments in offline RL, several researchers in ride-sourcing have proposed adopting vanilla offline RL methods or off-policy RL techniques such as Q-learning and SARSA \cite{sutton1998introduction} to learn value functions from batches of past vehicle transitions. Specifically, \cite{jiao2021real} employs a batch training algorithm through TD learning with deep value networks to learn the spatiotemporal state-value function from historical data, optimizing repositioning actions in ride-hailing services. \cite{yu2022batch} introduces Fitted Q-Iteration for optimizing vacant taxi routing, leveraging archived GPS data from taxis to train models without the need for real-time environment interaction. Additionally, the framework proposed in \cite{wei2023reinforcement} combines linear programming (LP) and tabular TD learning to dynamically reposition idle vehicles in ride-hailing systems. Here, LP manages the immediate repositioning based on a T-step lookahead prediction, while tabular TD learning is utilized to optimize long-term outcomes by learning from historical data and adjusting policies accordingly. This dual approach ensures both immediate responsiveness and strategic foresight in vehicle repositioning. Although these approaches significantly reduce computational costs and training time compared to traditional online Q-learning methods in ride-sourcing scenarios, adopting off-policy RL algorithms can be challenging due to extrapolation errors stemming from the OOD shift between the policy that collected the data and the learned policy \cite{kumar2020conservative}. This issue is particularly pronounced in increasingly complex ride-pooling systems where batches of transitions are likely to be incomplete for the large state-action decision space. Moreover, none of the cited works have explored the problem of offline RL policy fine-tuning in complex ride-pooling systems.

\subsection{Recent Advances and Fine-tuning Dilemma in Offline RL}\label{subsec_Offfline_RL_Review}
Recent progress in offline RL domain is mainly based on {conservatism(or pessimism)}~\cite{levine2020offline}. One direction is to enforce the learned policy to stay close to the behavior policy e.g. by applying Lagrangian through advantage weight regression \cite{peng2019advantage}, or by adopting TD learning with behavior cloning and constraint \cite{fujimoto2019off,fujimoto2021minimalist}. To exemplify, \cite{fujimoto2019off} employs a generative model conditioned on the state to produce actions that have been previously observed in the dataset, avoiding actions outside this distribution. This model is integrated with a Q-network that selects the highest valued action among those generated, ensuring that the actions are both plausible according to the data and valuable in terms of expected reward.  \cite{peng2019advantage} proposes an off-policy reinforcement learning algorithm that relies on simple supervised regression steps to optimize policy and value functions using previously collected data from a replay buffer. However this direction requires user to have pre-knowledge or estimation of sample policy for aggregated data, which might not be feasible in many real-world scenarios. Another direction is to implement critic regularization to the value function update like Conservative Q-learning \cite{kumar2020conservative} and Implicit Q-learning \cite{kostrikov2021offline} to directly address extrapolation error. Specifically, \cite{kumar2020conservative} intentionally lower-bounds the expected value of a policy through a regularization term added to the Q-function update, which penalizes overestimation of action values under the policy being evaluated. \cite{kostrikov2021offline} avoids directly querying the value of unseen actions by utilizing state value functions treated as random variables dependent on actions. The approach alternates between fitting an upper expectile of this random variable and updating a Q-function through SARSA-style temporal difference backup. Building on this trend, researchers have further showcased the capabilities of offline RL  to learn from human demonstrations in scenarios, such as robotics \cite{chebotar2023q} and multi-task learning in Atari games \cite{kumar2022offline}. This progress has been achieved by incorporating advanced neural network architectures, specifically the ResNet \cite{he2016deep} and Transformer \cite{vaswani2017attention} into these systems respectively. 

Although these exciting progress turns out to become effective for addressing extrapolation error, researcher find that there always exists a gap between offline learning and online learning. This gap results in current offline RL methods suffering from slow learning or initial unlearning during online further training \cite{nakamoto2024cal}. To bridge this gap, several branch of efforts have been made in the offline RL community. The first approach, as proposed by \cite{song2022hybrid}, involves loading pre-collected data trajectories as an initial experience batch for off-policy RL methods and then conducting training as usual. While this method is simple and efficient, its performance is still upper-bounded by state-of-the-art pure offline RL pipelines. Another line of approach, discussed by \cite{lee2022offline, mark2022fine}, involves training multiple value functions pessimistically during the offline training process. Afterwards during online fine-tuning, agent then performs actions based on the expected estimation of the ensemble value function. Though the ensemble value function manages to further reduce OOD overestimation,  the method could become memory inefficient due to the requirements for training over tens of neural networks simultaneously. The final approach aims to lower-bound the offline training of a conservative Q-function with a pre-trained reference value function \cite{nakamoto2024cal}. However, because the lower-bound of the pre-trained reference value function is crucial for the proposed framework and how inaccurate reference value function will impact the performance of the policy remains unknown, this method requires prior knowledge of the sample policy or environment system. The challenges above limit the adoption of these popular offline RL methods in our complex problem like coordinating ride-pooling with public transit.

With the rise and popularity of LLM like GPT-4, designing neural network and learning reward model or estimator to facilitate RL fine-tuning according to pre-collected human feedback has gained more attention than ever before. Specifically, \cite{ziegler2019fine} and \cite{ouyang2022training} proposed training reward models based on human preferences with supervised learning to better facilitate the training of RL. As an extension, \cite{rafailov2024direct} attempts to encode the reward model directly into the update of the RL policy. However, none of these ideas have been applied to the specific context of multimodal transportation systems. Inspired by the aforementioned works, we propose RG-CQL, where we managed to train a Guider during offline RL stage for reward regression and utilize the Guider to guide the exploration during online fine-tuning of RL policy. Under our novel offline training and online fine-tuning RL framework, RG-CQL could learn both efficiently and effectively for the complex task of coordinating ride-pooling with transit.

\section{Problem Description}\label{sec_prosetup}

\begin{figure}
    \centering
    \includegraphics[width=0.8\linewidth]{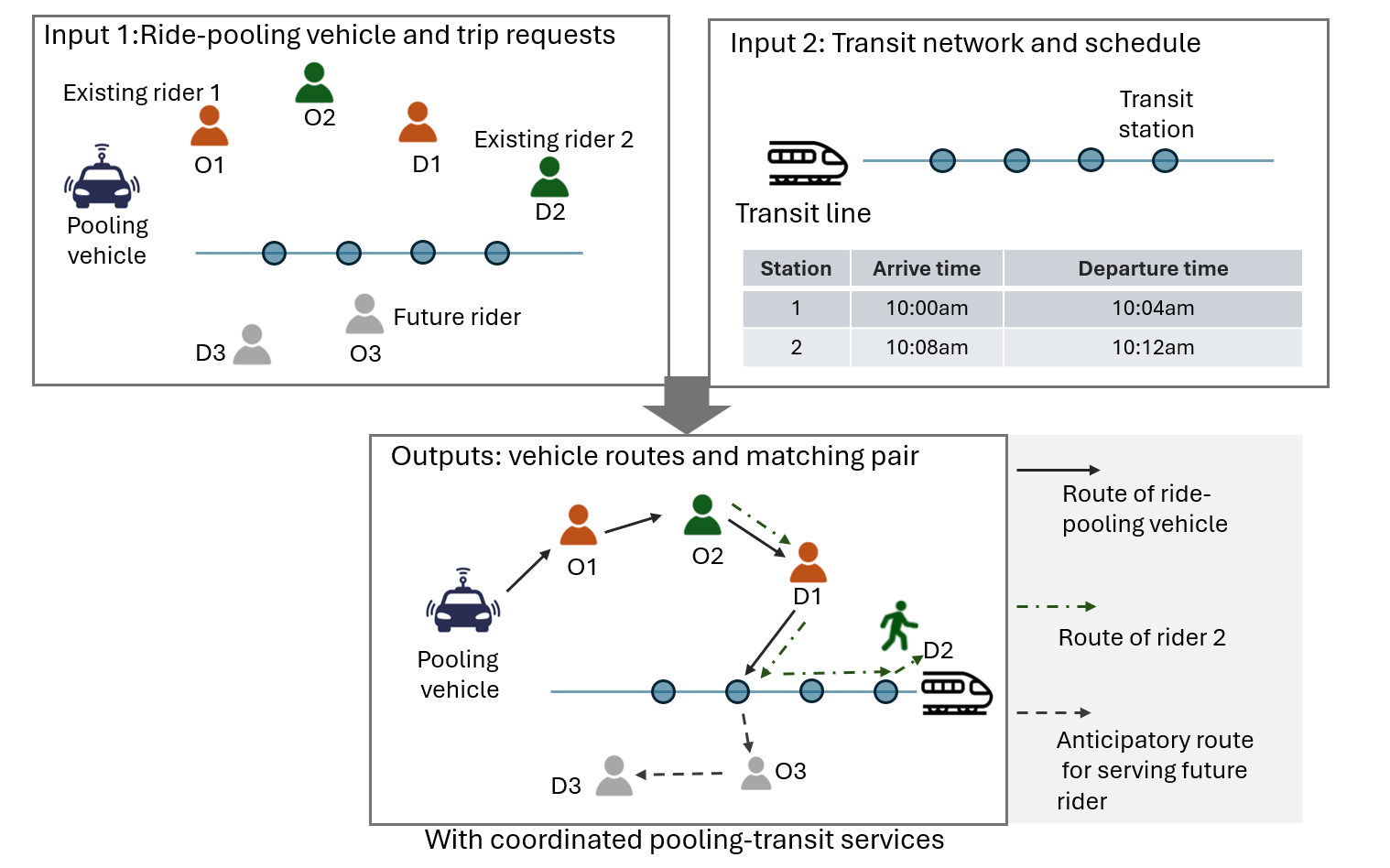}
    \caption{Problem setup with coordinated ride-pooling and transit services (Part of icons are from~\cite{GoogleImages})}
    \label{fig_problemsetup}
\end{figure}

We consider using ride-pooling services to address the first-mile problem of transit services. Riders requesting trip services are assumed to be willing to use ride-pooling services on their first leg of trips and transit services on the second leg of their trips. As shown in Fig.~\ref{fig_problemsetup}, a ride-pooling vehicle could pick up a rider from her origin and deliver the rider directly to her destination or to a transit station so that the rider can take transit vehicles. Riders who continue their trip using transit would get off at the stations closest to their destinations that are accessible via walking. The transit network comprises a collection of transit stations $\mathcal{I}$, along with transit routes linking these stations. Transit vehicles are assumed to follow fixed routes and schedules that are not influenced by ride-pooling services. The schedules of transit vehicles are given by a timetable that specifies a transit vehicle's arrival and departure times at each transit station $i\in \mathcal{I}$. Riders dropped at transit stations are expected to follow the shortest transit route to minimize travel time to their destinations, potentially transferring between transit lines.  Given the transit network and schedules, a centralized platform dispatches ride-pooling vehicles to riders and decides whether to drop off riders at their destinations or transit stations. For differentiation, we refer to a match as a "pooling-only" match if a rider is directly delivered to her destination and as a "pooling-transit" match if a rider is dropped off at a transit station and uses both ride-pooling and transit services to complete her trip. During a rider's journey on a ride-pooling vehicle, she might share a ride with other riders, depending on the route of the ride-pooling vehicle and the itinerary of other riders. 

We consider the operation of the above coordinated ride-pooling and transit services over a finite planning horizon $\mathcal{T}=\{0,\Delta t, 2\Delta t,\cdots, T\}$ where $\Delta t$ denotes time-step for decision making. The ride-pooling requests are revealed over time and arrive following a stochastic process. The platform is supposed to use a batch-matching strategy that matches a set of ride-pooling vehicles $\mathcal{N}$ with riders $\mathcal{M}$ that accumulated over a fixed time interval $\Delta t$ (such as 1 minute). Each rider $m\in \mathcal{M}$ is characterized by parameters $(o_m,d_m,e_m,l_m)$ which specifies rider $m$'s origin location $o_m$, destination location $d_m$, service request time $e_m$, and the expected arrival time $l_m$. The trip delay of a rider is calculated as the difference between a rider's actual arrival time at her destination and her expected arrival time, including the delay incurred by ride-pooling services and by transit services (if a match is a "pooling-transit" match). At time $t$, each ride-pooling vehicle $n\in \mathcal{N}$ is characterized by its current location $l_{n,t}$, remaining capacity $v_{n,t}$ to accommodate additional riders, riders that are being delivered $p_{n,t}$, and a scheduled route for picking up and dropping off those assigned riders in $p_{n,t}$. At each decision time, the platform anticipates future potential trip demand and maximizes its expected profits by deciding (i) the matching between ride-pooling vehicles $n\in \mathcal{N}$ and riders $m\in \mathcal{M}$; (ii) the drop-off locations (transit stations $i\in \mathcal{I}$ for a "pooling-transit" match or true destination for a "pooling-only" match) for each assigned rider $m\in \mathcal{M}$; (iii) the route and schedules that a ride-pooling vehicle follows to serve assigned riders. The matching decisions under the coordinated ride-pooling and transit services involve ride-pooling vehicles, riders, and transit stations. These decisions are expected to satisfy a set of service requirements, ensuring that riders can arrive at their destinations as soon as possible. Throughout the analysis, we assume that a ride-pooling vehicle is assigned to serve at most one rider at each decision time. This assumption does not preclude pooling possibilities among riders as we consider a dynamic system and riders assigned at present could share a ride with those assigned before. The overall goal is to efficiently link individual riders to public transit stations via ride-pooling services, thus facilitating riders' subsequent travel to final destinations while proactively managing future ride requests.

\section{Model Formulation}\label{sec_formulation}

\subsection{MDP Model}
This subsection introduces the MDP model for coordinated ride-pooling and transit services with each ride-pooling vehicle being treated as an agent. 

\subsubsection{Model Elements}
To streamline the model formulation, we partition the study area into a set of disjoint zones denoted by $\mathcal{Z} = \{1, \dots, z, \dots\}$. Within each zone $z \in \mathcal{Z}$, the transit stations are represented as $I_z$. For any rider dropped off in zone $z$, her specific drop-off transit station, which optimizes travel time to her final destination $d_m$, can be uniquely determined given the set $I_z$. Consequently, the drop-off location of a rider $m$ is represented either by $0$ (indicating direct drop-off at her destination) or by the zone number $z$. Building on this framework, the subsequent discussion elaborates the model components in detail, including the state $S_t$, action $A_t$, rewards $R_t$, transition function $P$, and the discount factor $\gamma$.


{
\setlength{\parindent}{0pt}
1) {\textbf{State}}: At time $t\in \mathcal{T}$, we denote the state of a ride-pooling vehicle $n\in \mathcal{N}$ by a tuple $s_{n,t}=(t,l_{n,t},v_{n,t}, p_{n,t}, o_{m}, d_{m})$, which encapsulates the current time $t$, the vehicle's location $l_{n,t}\in \mathcal{Z}$, the remaining vacant seats $v_{n,t}$, the origin $o_m$ and destination $d_m$ of a matched rider $m$ awaiting pick-up at their origin, and the information of riders on board $p_{n,t}$. To facilitate modeling, we introduce a dummy zone indexed by $z'$. When the ride-pooling vehicle $n$ is fully occupied, i.e., $v_{n,t}=0$, $o_m$ (or $d_m$) takes the value $z'$, indicating the vehicle's unavailability for matching. Otherwise, $o_m\in \mathcal{Z}$ and $d_m\in \mathcal{Z}$. The vector $p_{n,t}$ accommodates scenarios where the ride-pooling vehicle $n$ may be currently on its way to pick up assigned riders or transporting riders (we call these prior assigned riders as passengers thereafter) to their destinations or intermediate transit stations. This vector is represented as $p_{n,t}=(i_1,i_2, \cdots i_{c_m},t_{1},t_{2}, \cdots t_{c_m}, \delta t_{1},\delta t_{2},\cdots \delta t_{c_m})$, where $i_k\in \mathcal{Z}\cup {0}\cup z'$ denotes the $k$th passenger's drop-off location, $t_{k}$ is the $k$th passenger's estimated remaining time on board, and $\delta t_{k}$ represents $k$th passenger's additional travel time as opposed to non-pooling services, where passengers are delivered directly from their origins to their destinations without sharing a vehicle. In cases where the ride-pooling vehicle $n$ is partially occupied, i.e., $0<v_{n,t}<c_m$, we set $i_{k}=z'$, $t_{k}=0$, and $\delta t_{k}=0$ for $\forall k\in (c_m-v_{n,t},c_m]$. The values of $v_{n,t}$ and $p_{n,t}$ essentially convey the occupancy status and routing information of the ride-pooling vehicle $n$. At time $t$, the system state $S_t$ aggregates the states of ride-pooling vehicles and is denoted as $S_t= [{s_{1,t}},\ldots ,{s_{n,t}}]$.


2) {\textbf{Action}}: For a fully occupied ride-pooling vehicle where $v_{n,t}=0$, the only available action is to continue its route for picking up and dropping off riders. In the case of a partially occupied ride-pooling vehicle, where $v_{n,t}>0$, the vehicle can take the following two types of actions upon receiving a trip request from rider $m$: it either drops off rider $m$ at her final destination or at an intermediate transit station $i \in \mathcal{I}$, allowing the rider to transfer to transit for the remaining journey. For both types of actions, we suppose that a ride-pooling vehicle follows the route of the shortest travel distance to collect and deliver the rider to the designated drop-off locations. Also, recall that a rider's route on transit is uniquely determined given the zone where she is dropped off. Consequently, at time $t$, the ride-pooling vehicle $n$'s route for accommodating a new rider $m$ can be uniquely determined based on the rider's drop-off zone and the information regarding previously assigned passengers $p_{n,t}$. Given this context, seeing rider $m$, the action space of a ride-pooling vehicle is represented by $0\cup \mathcal{Z}$, where action $a_{n,t}=0$ and $a_{n,t}=z$ signify that vehicle $n$ drops off a rider at her final destination or a station located in zone $z\in \mathcal{Z}$, respectively. At time $t$, the collection of all agents' actions is denoted as ${A_t} = [{a_{1,t}},{a_{2,t}}, \ldots ,{a_{N,t}}]$. 


{To enhance the explanation of our action and state formulation in the MDP model, we present an intuitive example based on the problem setup previously illustrated in Figure~\ref{fig_problemsetup}. Consider the scenario depicted in Figure~\ref{fig_MDP}: at time $t$, vehicle agent $n$, with a seating capacity of three, is located in zone 1, having just picked up rider 1, marked in orange. Here, $l_{n,t} = 1$ indicates the vehicle's current location, and $v_{n,t} = 2$ shows that there are two vacant seats. The passenger information vector $p_{n,t} = (3,0,0,8,0,0,0,0,0)$, where the first element `3' specifies the destination zone of rider 1 and `8' denotes his remaining time onboard in minutes. Given that rider 1 is traveling without recourse to public transit and no other passengers are sharing the ride, no additional detour is necessary, reflected by the zero in the third last position $\delta t_{1}$ of $p_{n,t}$. The remaining elements of $p_{n,t}$ are set to zero, indicating that no other passengers are onboard. Following this, rider 2, shown in green and originating from zone 2 ($o_m = 2$), is destined for zone 5 ($d_m = 5$). The state of vehicle agent $n$ is then represented as $s_{n,t} = (t, l_{n,t}, v_{n,t}, p_{n,t}, o_m, d_m)$. An action $a_{n,t} = 4$ is assigned to drop off rider 2 at zone 4, prompting a reroute (indicated by the black arrows in Figure~\ref{fig_MDP}), which incurs an additional one-minute detour for rider 1, calculated as ($4 + 5 - 8 = 1$ min), due to a deviation from their initial direct route, updating rider 1's remaining onboard time to 9 minutes. For rider 2, compared to a direct route (illustrated in green), a six-minute extra detour ($5 + 6 + 6 + 1 - 12 = 6$ min) arises from the integration of ride-pooling and transit, extending rider 2's remaining time onboard to 11 minutes. Consequently, the passenger information updates to $p_{n,t+1} = (3,4,0,9,11,0,1,6,0)$, where `3' and `4' denote the destination zones of riders 1 and 2, respectively; `9' and `11' represent the respective updated remaining times onboard; and `1' and `6' reflect the additional detours for riders 1 and 2, respectively. The remaining elements are set to zero, reflecting the vacancy of the third seat.}

\begin{figure}
    \centering
    \includegraphics[width=0.8\linewidth]{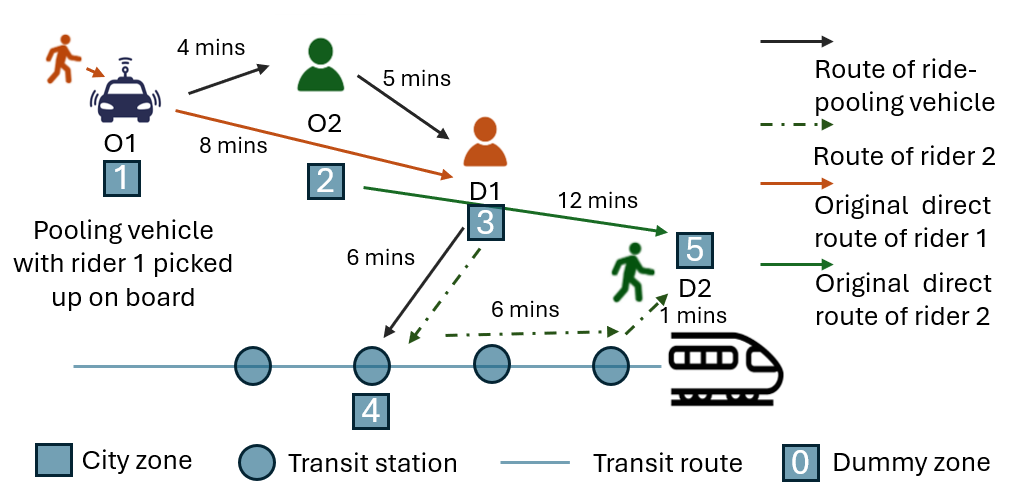}
    \caption{{Intuitive example of MDP formulation. Consider at time $t$, agent $n$ just picked up rider 1 and observe matched rider 2, $l_{n,t}= 1, v_{n,t}= 2, o_{m}= 2, d_{m}= 5, p_{n,t}= (3,0,0,8,0,0,0,0,0)$, where the first element `3' specifies the destination zone of rider 1 and `8' denotes his remaining time onboard in minutes. The vehicle agent is assigned an action $a_{n,t} = 4$ to drop off the passenger at zone 4, allowing rider 2 to continue his/her journey via public transit. After the match, with the update of new route (in black arrows), rider 1 experiences 1 minute additional detour and remaining onboard time is updated to 9 minute. Rider 2 experiences 6 minute detour due to the combined usage of ride-pooling and transit and the remaining onboard time is 11 minutes. Consequently, the passenger information on board will update to $p_{n,t+1} = (3,4,0,9,11,0,1,6,0)$, where `3' and `4' denote the destination zones of riders 1 and 2, respectively; `9' and `11' represent the respective updated remaining times onboard; and '1' and `6' reflect the additional detours for riders 1 and 2, respectively. The remaining elements are set to zero, reflecting the vacancy of the third seat.}}
    \label{fig_MDP}
\end{figure}

3) {\textbf{Reward}}: We denote $r_{n,t}$ as a ride-pooling vehicle $n$'s reward for taking action $a_{n,t}$ at state $s_{n,t}$. For a fully occupied ride-pooling vehicle ($v_{n,t}=0$), its reward for continuing to serve its passengers is set as zero. For a partially occupied ride-pooling vehicle assigned to serve new rider $m$, its rewards for taking action $a_{n,t}$ at state $s_n$ is equal to the revenue for serving rider $m$ minus the increased time costs that both the new rider $m$ and existing passengers endure:
\begin{equation}\label{eq_reward}
\begin{aligned}
r_{n,t}(s_{n,t},a_{n,t}) &= \beta_{0} + \beta_{1} \cdot dis(o_m,d_m) -\beta_{2}\cdot w_{m} \\
                                &- \left (\beta_{3} \cdot \min \{\sum_{k:k\in m\cup \Omega_{n,t}} {\delta t_k},\kappa\}+\beta_{4} \cdot \max \{ \sum_{k:k\in m\cup \Omega_{n,t}} {\delta t_k} -\kappa, 0\}\right) \\
                                &+\left(\beta_{3} \cdot \min \{\sum_{k:k\in \Omega_{n,t}} \delta t_k', \kappa\}+\beta_{4} \cdot \max \{ \sum_{k:k\in \Omega_{n,t}} \delta t_k'-\kappa, 0\}\right),
\end{aligned}
\end{equation}
where parameter $\beta_1$ denotes the cost per unit distance; parameters $\beta_2,\beta_3,\beta_4$ represent the costs per unit time; $t_k'$ and $t_k$ represent the additional detour time for passenger $k$ before and after the execution of action $a_{n,t}$, respectively. The first term $\beta_0$ on the right-hand side of the above equation represents flag fare for serving a new rider. The second term prescribes the distance-based revenue for serving a new rider $m$, which is assumed to only depend on the {spatial (Euclidean)} distance\footnote{{The distance $dis(o_m,d_m)$ is computed using the Euclidean distance between points. It is noteworthy that this simplification is unlikely to significantly influence our findings, as prior studies have demonstrated that scaling the Euclidean distance by a constant adequately approximates the street distance. i.e., the distance of the shortest path over road networks}~\cite{boscoe2012nationwide,boyaci2021vehicle}. } $dis(o_m,d_m)$ between the rider's origin $o_m$ and destination $d_m$. The third term $\beta_{2}\cdot w_{m}$ denotes the waiting time costs of rider $m$, and $w_{m}$ represents rider $m$'s waiting time at her origin before she is picked up. The second and the third lines on the right-hand side of the equation compute the increased trip delay cost after action $a_{n,t}$ is executed, with the trip delay costs before action $a_{n,t}$ is executed serving as the baseline. The bracket expression within the second line computes the total additional travel time of riders, including the new rider $m$ and existing passengers (denoted by set $\Omega_{n,t}$), for using ride-pooling services after action $a_{n,t}$ is executed. Here, we penalize the total additional travel time of served riders using a piecewise linear function with $\kappa$ being a delay threshold. When riders' total additional travel time $\sum_{k:k\in m\cup \Omega_{n,t}} {\delta t_k}$ is within the delay tolerance $\kappa$, the trip delay cost is calculated $\beta_{3} \cdot \min \{\sum_{k:k\in m\cup \Omega_{n,t}} {\delta t_k},\kappa\}$. Otherwise, a larger penalty $\beta_{4}>\beta_{3}$ is introduced to further penalize the part of additional travel time larger than $\kappa$, leading to the extra term $\beta_{4} \cdot \min \{\sum_{k:k\in m\cup \Omega_{n,t}} {\delta t_k}-\kappa,0\}$. On the third line of the equation, the same rule is used to calculate the total trip delay costs for those existing passengers before action $a_{n,t}$ is executed. Given $r_{n,t}$, the profit of the system at time $t$, denoted as $R_t(S_{t},A_{t})$, sums the rewards of all agents, i.e., ${R_t}(S_{t},A_{t}) = \sum_{n:n\in \mathcal{N}} r_{n,t}(s_{n,t},a_{n,t})$. 


4) {\textbf{Transition Function}}: The transition function is denoted as \(P(S_{t+1}|S_t, A_t)\), which encapsulates the probabilities of transitioning from a current state \(S_t\) to a state \( S_{t+1} \) at time $t+1$ contingent upon the execution of action \(A_t\). The transition probabilities $P(S_{t+1}|S_t, A_t)$ depend on the occupancy states of ride-pooling vehicles. In case a fully occupied ride-pooling vehicle $n$ does not drop off any passengers during time interval $(t,t+1]$, its states at time $t+1$ is determined, leading to a transition probability of either 0 or 1 from state $s_{n,t}$ to a state $s_{n,t+1}$. For other ride-pooling vehicles, their probabilities transiting from  $s_{n,t}$ to $s_{n,t+1}$ are influenced by the platform's matching decisions at time $t$ and external arrival of riders' requests. In our study, both the reward function and transition probability do not need to be explicitly modeled, and they will be learned by the RL algorithm. 

5) {\textbf{Discount Factor}}: The discount factor $\gamma$ quantifies the present value of future rewards, whose value lies in the interval $[0,1]$.


\subsubsection{Policy and Platform Objective}

Let $\Pi(A_t|S_t)$ be a centralized policy mapping system states $S_{t}$ to actions $A_t$, which defines a distribution over actions given states. Let $Q_{\Pi}(S_t, A_t)$ denote the platform's expected return starting at state $S_t$, taking action $A_t$, and then following policy $\Pi$. We compute $Q_{\Pi}(S_t, A_t)$ by the following formula:
\begin{equation}
    Q_{\Pi}(S_t, A_t)=
    \begin{cases}
        & R_t(S_t,A_t), \quad \text{if $t=T$,}\\
        &\sum_{S_{t+1}} P(S_{t+1}|S_t,A_t)\cdot (R_t(S_t,A_t)+\gamma V_{\Pi} (S_{t+1})), \quad \text{Otherwise.}\\
    \end{cases}
\end{equation}
where $V_{\Pi}(S_t)$ be the platform's expected return starting from state $S_t$ and following policy $\Pi$, i.e., $V_{\Pi}=0$ if $t=T$ and $V_{\Pi}(S_t)=\sum_{A_{t}} \Pi(A_t|S_t) Q_{\Pi}(S_t, A_t)$ otherwise. Essentially, $Q_{\Pi}(S_t, A_t)$ computes the expected cumulative rewards of the platform starting at state $S_t$, taking action $A_t$, and then following policy $\Pi$:  
\begin{equation}
\begin{aligned}\label{eq:orignal Qtotal}
 Q_{\Pi}(S_t, A_t) = \mathbb{E}_{\Pi} \left[ \sum_{\tau:\tau\in K_t} \gamma^\tau \cdot R_{t + \tau + 1} \mid S_t, A_t \right],
\end{aligned}
\end{equation}
where $K_t=\{0,1,2,T-t\}$ denotes the set of time steps afterwards $t$ until the end of planning horizon. The objective of the platform is to determine the optimal policy $\Pi^{*}$ that maximizes the platform's expected discounted cumulative reward over the whole planning horizon. 

Directly solving policy $\Pi$ presents significant challenges due to the curse of dimensionality. Note that the platform's expected return depends on how the system states evolve, which further relies on the actions of all agents. Also, at each decision time, thousands of agents might need to be simultaneously dispatched, leading to a prohibitively high dimensional state and action space. To simplify analysis and facilitate the model solution, we make the following two assumptions, which are very common in the literature of multi-agent RL \cite{al2019deeppool,singh2021distributed,tang2021value,sadeghi2022reinforcement, feng2022coordinating, wang2023optimization}: (i) agents are independent in that an agent's reward and state transition probability only depends on its own state and actions, which is independent of the states and actions of other agents; (ii) agents are homogeneous and thus share the same policy $\pi(a_{n,t}|s_{n,t})$. Under the independence assumption, the dimensionality of the MDP model can be reduced by decentralizing the state transition function. Specifically, the transition function $P(S_{t+1}|S_{t},A_{t})$ can be expressed as:
\begin{equation}
    P(S_{t+1}|S_{t},A_{t})=\prod_{n:n\in \mathcal{N}} p(s_{n,t+1}|s_{n,t},a_{n,t})
\end{equation} where $p(s_{n,t+1}|s_{n,t},a_{n,t})$ specifies an agent's probability transiting from state $s_{n,t}$ to state $s_{n,t+1}$ contingent on taking action $a_{n,t}$. Consequently, the platform's expected return $Q_{\Pi}(S_t, A_t)$ can be calculated as the sum of all agents' expected return:
\begin{equation}
    Q_{\Pi}(S_t, A_t)= \sum_{n:n\in \mathcal{N}} Q_{\Pi}(s_{n,t},a_{n,t})
\end{equation}
The second assumption states that ride-pooling vehicles are homogeneous, which is a reasonable assumption in the context of ride-pooling fleet because all the vehicles are assumed to have the identical characteristics (e.g., seat capacity). This assumption further transfers the platform's expected return into:
\begin{equation}
    Q_{\pi}(S_t, A_t)= \sum_{n:n\in \mathcal{N}} Q_{\pi}(s_{n,t},a_{n,t})
\end{equation}
where $\pi$ is the policy for each individual agents, that are shared by all agents on the platform. 

\begin{remark}
{We clarify that the assumption on independence holds because competition among agents in the context considered in this paper is actually not strong enough to necessitate the explicit consideration of interdependence. Intuitively, the competition and correlation among each agent (i.e., vehicle) is stronger in the case of low demand and high supply. This is because, in such scenarios, vehicles face fiercer competition with neighboring vehicles for passengers. On the other hand, if the supply is limited but the demand is very high, then the competition among vehicles would be much weaker, since even if a vehicle cannot be assigned a passenger due to the presence of another vehicle, it has a much larger chance of being assigned to another passenger due to the abundance of demand. This indicates that the presence of competitors in this case will have a significantly smaller impact on the ego agent since it has a much smaller chance of being unassigned in the end due to the competition. Fortunately, our model is specifically tailored for peak hours where the demand (number of orders) exceeds the supply capabilities of the vehicle agents, because in this case the platform has stronger incentives to integrate public transit with ride-pooling. In these scenarios, the competitive element among agents is naturally mitigated, as the focus shifts towards leveraging public transit resources to increase the chance of ride-pooling and fulfill excess demand.} 
\end{remark}

With the above two assumptions, optimizing policy $\Pi$ is equivalent to determining a policy $\pi^{*}$ that maximizes the platform's expected gains over the whole planning horizon, i.e.,
\begin{align}
    \pi^{*}=\argmax_{\pi} {\mathbb{E_{\pi}} \left[ \sum_{n:n\in \mathcal{N}} \sum_{\tau:\tau\in K_t} \gamma^\tau \cdot r_{n,t + \tau} \mid s_{n,t}\right]}.
\end{align}
where $s_{n,t}$ denotes the current states of ride-pooling vehicle $n$ $\forall n\in \mathcal{N}$.

\subsection{Real-time Order Dispatch and Drop-off Location Choices}
In this subsection, we formulate a bipartite matching problem for real-time order dispatching decisions. The optimization problem outputs the matching pairs between ride-pooling vehicles and waiting riders, along with the corresponding drop-off locations for assigned riders. At each decision time $t$, this problem is solved to guide the actions of each agent $a_{n,t}$ based on system state $S_t$ so that the platform's expected return is maximized.  The value for each vehicle-rider match is determined by RL considering the exploration and exploitation trade-off, which is integrated into the optimization problem's objective function.
\begin{figure}[!htbp]
  \centering
  \includegraphics[width=0.67\linewidth]{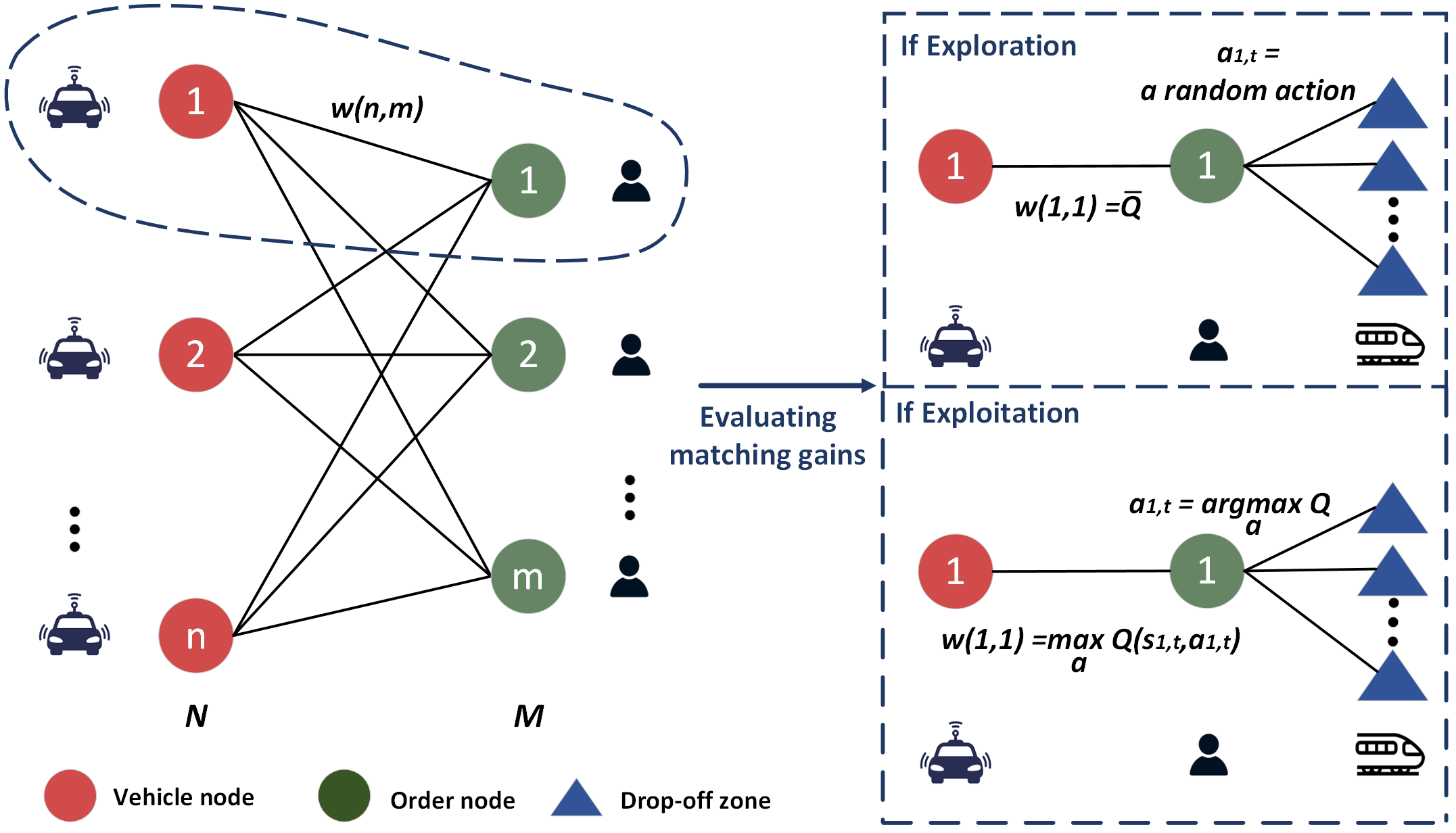}
  \caption{Visualization of bipartite matching}
  \label{fig:BM}
\end{figure}

We introduce the bipartite graph, based on which the optimization problem is formulated. As shown in Fig. \ref{fig:BM}, we represent available ride-pooling vehicles and riders waiting to be matched as two sets of nodes $\mathcal{N}$ and $\mathcal{M}$, respectively. Edges connect each vehicle node $n$ in set $\mathcal{N}$ to each rider node $m$ in set $\mathcal{M}$. A weight $w(n,m)$ is associated with the edge connecting nodes $n$ and $m$, which measures the gains for matching ride-pooling vehicle $n$ with rider $m$. The value of weights $w(n,m)$ depends on which action $a_{n,t}$ that an agent $n$ takes or more specifically depends on the drop-off locations of rider $m$. 
To make decisions efficient in the long run, we determine the value of weight $w(n,m)$ and action $a_{n,t}$ taking into account the future system states. Inspired by \cite{sutton1998introduction}, we use an $\epsilon$-greedy strategy that leverages exploitation and exploration to address the long-term uncertainties inherent in the ride-pooling services. Specifically, for exploitation, we select the action $a_{n,t}$ that maximizes the agent's expected return $Q_{\pi}(s_{n,t},a_{n,t})$ for given agent $n$ at state $s_{n,t}$. Correspondingly, weight $w(n,m)$ takes value $\max_{a_{n,t}:a_{n,t}\in \mathcal{Z}\cup 0} Q_{\pi}(s_{n,t},a_{n,t})$. For exploration, the agent $n$ is assigned a random action $a_{n,t}\in \mathcal{Z}\cup 0$, meaning that the ride-pooling vehicle $n$ would drop off a rider $m$ at a randomly selected zone. The corresponding value of $w(n,m)$ is set as a large positive number $\overline{Q}$, driving the agent to take such a random action after being matched. The trade-off between exploitation and exploration is controlled by a parameter $\epsilon \in (0,1)$ that specifies the exploration rate, and the corresponding weight values can be expressed as:  
\begin{equation} \label{eq:Matching Value Function} 
\begin{cases} 
& w(n,m)= \underset{a_{n,t}} \max \, Q_{\pi}(s_{n,t}, a_{n,t}), \quad  \text{with probability } 1 -\epsilon  \\
& w(n,m)=\overline{Q}, \quad \text{with probability } \epsilon 
\end{cases}
\end{equation}

With the bipartite graph defined above, the action of agents $a_{n,t}$ and the resultant expected returns can be uniquely determined at a decision time $t$ once the selected edges linking vehicle nodes and rider nodes are established. To this end, we introduce a variable $x_{n,m}$ for each edge, which equals 1 if the edge connecting nodes $m$ and $n$ is selected and 0 otherwise. The following ILP program is formulated for decisions on order-dispatching and drop-off locations, which merges reinforcement learning's policy function with a bipartite matching process: 
\begin{subequations} \label{eq:ILP} 
\begin{align} 
& \underset{X} {\text{max}} 
\sum_{n:n\in \mathcal{N}} w(n,m) \cdot x_{n,m}, \label{match_obj}\\ 
\text{s.t.} \quad 
& \sum_{n:n\in \mathcal{N}} x_{n,m} \leq 1, \quad \forall m\in \mathcal{M},\label{match_order} \\ 
& \sum_{m:m\in \mathcal{M}} x_{n,m} \leq 1, \quad \forall n\in \mathcal{N}, \label{match_driver}\\ 
& \sum_{n:n\in \mathcal{N}} x_{n,m}\cdot d_{n,m} \leq R_{match}, \quad \forall m\in \mathcal{M}, \label{match_distance}\\
& x_{n,m} \in \{0,1\}, \quad \forall n\in \mathcal{N},m\in \mathcal{M},
\end{align} 
\end{subequations}
where $X=\{x_{n,m}\}_{n\in \mathcal{N},m\in \mathcal{M}}$ denotes the set of decisions variables. The objective \eqref{match_obj} is to maximize the platform's expected profits. Constraint \eqref{match_order} ensures that each order can be matched with at most one ride-pooling vehicle while Constraint \eqref{match_driver} guarantees that each ride-pooling vehicle is matched with at most one order at the decision time. Constraint \eqref{match_distance} guarantees that a ride-pooling vehicle and an order can be matched only if the distance between them $d_{n,m}$ is within a maximum matching distance $R_{match}$.

\section{RG-CQL for Value Function and Policy Learning}\label{sec_RL}


This section introduces RG-CQL, an offline RL pretraining and reward-guided online RL fine-tuning framework, for solving the MDP model presented in the last section. As aforementioned, introducing the coordinated "pooling-transit" services greatly increases state and action spaces compared to implementing "pooling-only" services. Consequently, implementing traditional RL methods to learn from scratch through online iterative interactions with a manually crafted simulator would lead to low training efficiency and potential local optimal solutions. RG-CQL aims to overcome this bottleneck by leveraging a diverse set of data that broadly covers the real-world on-demand ride-pooling system and state transitions. In Section \ref{subsec_offline}, we discuss the offline phase RL method which learns the best policy in support from existing data. In Section \ref{subsec_online}, we innovatively introduced a module named "Guider", which learns to predict agents' rewards from data in the offline training stage and guides the exploration of CDDQN in the online fine-tuning/deployment stage. In Section \ref{subsec_overview}, we summarize our RL algorithm.

\subsection{CDDQN for Offline Learning from Pre-collected Transitions}\label{subsec_offline}

At the offline stage, we develop CDDQN based on the idea of conservatism~\cite{levine2020offline}, aiming for learning Q value from existing data. 

As the first step, a batch of observations $\mathcal{D}$ regarding state transition are obtained from existing on-demand ride data, with $\mathcal{D}$ containing a series of trajectories $\tau=(s, a, r, s')$. Here, $s$ is the state of a vehicle, $a$ is a vehicle's action, $r$ is the reward, and $s'$ the vehicle's state after taking action $a$ at state $s$. To enrich the training dataset, we include not only ride-pooling data but also data on non-pooling services where riders are served individually\footnote{{Specifically, under non-pooling scenarios, the state representation of vehicle agent could be taken as $s_{n,t,NP} = (t, l_{n,t}, v_{n,t,NP}, p_{n,t,NP}, o_m, d_m)$ with seat capacity as 1 and thus $p_{n,t,NP}=(i_1, t_{c_1}, \delta t_{1})$. To integrate non-pooling data into a format suitable for pooling scenarios, we could argument its state into $s_{n,t,P} = (t, l_{n,t}, v_{n,t,P}, p_{n,t,P}, o_m, d_m)$ with argument seat capacity as 3, and $v_{n,t,P} = v_{n,t,NP} + 2 $ and $p_{n,t,P} = (i_1, 0, 0, t_{c_1}, 0, 0, \delta t_{1}, 0, 0)$, where the other elements are set as `0' because of vacancy of the remaining seats. The other components of our model, such as the representations for reward and action, remain unchanged.}}. Despite this expansion, the definitions of vehicle state $s$, reward $r$, and action $a$ remain consistent with those outlined in coordinated ride-pooling and transit services. {In the real world, TNCs can effectively collect and organize data from past dispatch decisions as a feasible strategy. Specifically, under their current or previous policies (such as one or a combination of our four benchmark policies), TNCs can aggregate the initial status of each dispatched vehicle and the details of the matched orders to compile the state information. The destinations of the dispatches can then serve as the action data points. The rewards can be quantified by combining factors such as company revenue and passenger satisfaction, aligning with the reward function described in (\ref{eq_reward}).}



The second step approximates Q-value through CDDQN which has its roots in DDQN~\cite{van2016deep}. Given a batch of observations $\mathcal{D}$, the off-policy RL like DDQN updates Q-value by separating action selection from action evaluation. Initially, the algorithm identifies the optimal action to generate the TD target and subsequently assesses this action using a target network. The loss function $L$ of DDQN is defined as follows: 
\begin{equation} \label{eq:DDQN Loss Calculation}
\begin{aligned}
L = \mathbb{E}_{\tau \sim \mathcal{D}}\Bigg[\Bigg{(}r + \gamma Q\left(s',\mathop{\arg\max}_{a'} Q(s',a';\theta);\theta^- \right)
- Q(s,a;\theta)\Bigg{)}^2\Bigg], 
\end{aligned}
\end{equation}
where $Q(s,a;\theta)$ is the Q-value estimated by the training Q-network whose neural network parameter is $\theta$, and $Q(s,a;\theta^-)$ is the Q-value estimated by the target network $\theta^-$. Although DDQN is a straightforward method, directly applying it as fitted Q Iteration \cite{yu2022batch} in the offline training stage can result in significant extrapolation errors. This issue arises due to the pre-collected data batch covering only a fraction of the state-action space. Furthermore, including coordinated ride-pooling and transit services exacerbates this challenge by expanding the agents' action space, leading to unobserved states and actions within the existing dataset. To address the extrapolation error problem mentioned above, we leverage the concept of "conservatism" introduced in CQL \cite{kumar2020conservative}. Specifically, we add a conservative regularization term to Equation~\eqref{eq:DDQN Loss Calculation} and formulate the loss function $L_c$ of CDDQN as follows:
\begin{equation} \label{eq:CDDQN_Loss_Calculation}
\begin{aligned}
L_c &= \mathbb{E}_{\tau \sim \mathcal{D}}\Bigg[\Bigg{(}r + \gamma Q\left(s',\mathop{\arg\max}\limits_{a'} Q(s',a';\theta);\theta^- \right) - Q(s,a;\theta)\Bigg{)}^2\Bigg] \\
& + C \Bigg{(} \mathbb{E}_{s \sim \mathcal{D}}[\max\limits_{a} Q(s,a;\theta)] - \mathbb{E}_{(s,a) \sim \mathcal{D}}[Q(s,a;\theta)] \Bigg{)},
\end{aligned}
\end{equation}
where $C$ is a hyper-parameter requiring careful tuning and dictates the extent to which the regularization term $\Bigg{(} \mathbb{E}_{s \sim \mathcal{D}}[\max\limits_{a} Q(s,a;\theta)] - \mathbb{E}_{(s,a) \sim \mathcal{D}}[Q(s,a;\theta)] \Bigg{)}$ should be accounted for. This additional conservative regularization term penalizes Q-values associated with unobserved state-action pairs in the dataset, encourages the Q-values for unobserved state-action pairs to be minimized, particularly if these Q-values mistakenly emerge as the highest among all actions for a given state in the dataset. The purpose is to mitigate the risk of overestimating Q-values for state-action pairs not present in the dataset.

The network parameter updating process of CDDQN remains the same as that of DDQN. We adopt gradient descent to update the training network parameter $\theta$:
\begin{equation} \label{eq:CDDQN update}
\theta = \theta-\alpha_c\cdot {\nabla_\theta } L_c,
\end{equation}
where $\alpha_c$ is learning rate and ${\nabla _\theta } L_c$ denotes the gradient of CDDQN loss function $L_c$ with respect to parameters $\theta$. To stabilize the training process, we adopt Polyak Average for soft update \cite{fujimoto2018addressing} when updating the CDDQN target network, which maps training network parameters $\theta$ to target network parameters $\theta^-$ after every training step by the following formula: 
\begin{equation} \label{eq:Polyak Average}
\theta^- = \rho \cdot \theta + (1 - \rho) \cdot \theta^-,
\end{equation}
where $\rho$ is the soft update hyper-parameter.

\subsection{Reward Guided CDDQN for Online-finetuning}\label{subsec_online}

At the online stage, we deploy reinforcement learning algorithms developed during the offline stage to interact with real-world environments or simulations, aiming for further fine-tuning RL algorithms.


Fine-tuning CDDQN directly in an online setting can lead to solution inefficiencies, such as slow learning and initial unlearning, as highlighted in existing studies \cite{nakamoto2024cal}. This disparity arises from the divergence in Q-value estimation between offline and online learning stages. As indicated by Equation~\eqref{eq:CDDQN_Loss_Calculation}, the offline training stage diminishes Q-values associated with unobserved actions in the existing dataset. Conversely, during online learning, agents exhibit optimism towards unseen state-action pairs. This optimism is exemplified by the $\epsilon$-greedy strategy outlined in Equation~\eqref{eq:Matching Value Function}, where a large Q-value is allocated to a randomly selected action to foster exploration. This optimism-pessimism gap complicates the balance between offline RL and online fine-tuning, especially when agents explore the environment. Fig.~\ref{fig:RG-CQL}(a) illustrates the dilemma caused by such an optimism-pessimism gap between offline RL and online fine-tuning. An agent risks getting lost at the beginning of online fine-tuning if it indiscriminately explores all unseen state-action pairs and updates its strategy, potentially undermining the strengths of the original conservative offline RL policy.
\begin{figure}[ht]
  \centering
  \includegraphics[width=0.9\linewidth]{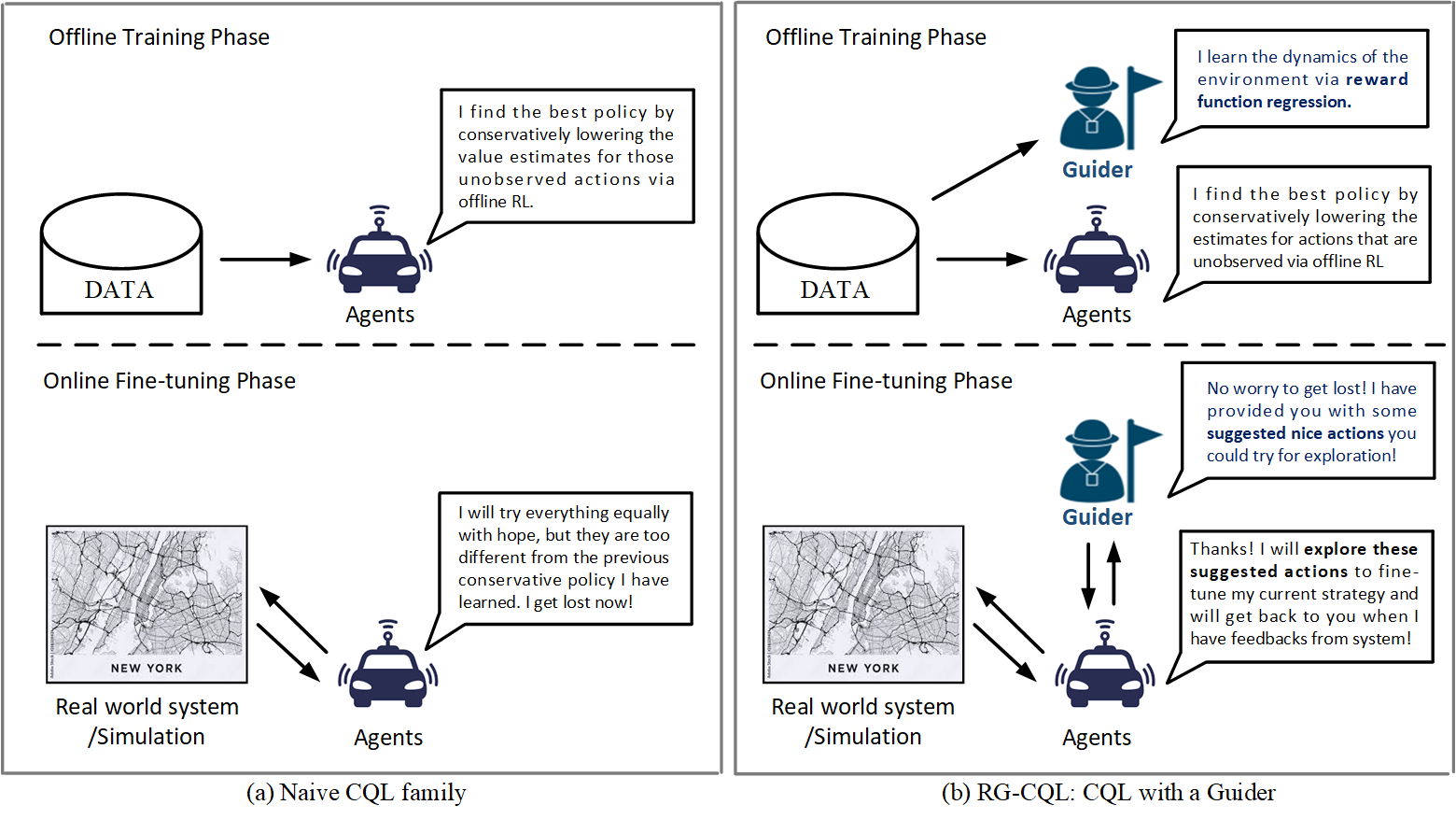}
  \caption{Comparison of naive CQL family pipeline (on the left) with RG-CQL (on the right) (part of icons from~\cite{GoogleImages})}
  \label{fig:RG-CQL}
\end{figure}

 Observing this, we introduce an innovative module, referred to as ``Guider", to resolve the issue caused by the pessimism-optimism gap. As illustrated in Fig.~\ref{fig:RG-CQL}(b), the Guider aims to enhance agents' exploration by suggesting less blindly optimistic actions while maintaining the potential to discover long-term optimal policies during the fine-tuning phase. To accomplish this, the Guider employs the reward function as the foundational model-based dynamics metric. It uses the instant reward $r$ to guide agents' action choices when exploration strategy is used in order-matching. The underlying premise is that state-action pairs generating significant negative short-term rewards, especially during peak demand periods, are likely to lead to a high number of rejected orders. Consequently, these pairs are unlikely to contribute to optimal long-term operational decisions, even when considering the trade-off between immediate rewards and long-term objectives. Therefore, by eliminating these unreasonable decisions from the exploration process, we can significantly guide agents to adopt a more conservative stance during online fine-tuning, while simultaneously improving the efficiency of exploration.
 
 To be more specific, during the offline stage, we train a Guider network using supervised learning in addition to the CDDQN training. The Guider network is a neural network that learns a function approximator for reward $r$ using existing data, which would be used in the online stage. The loss function for this Guider network is defined as follows: 
 \begin{equation} \label{eq:Reward Model Loss Calculation}
\begin{aligned}
L_g = \mathbb{E}_{\tau \sim \mathcal{D}}\Bigg[\Bigg{(}r - G(s,a;\phi)\Bigg{)}^2\Bigg], 
\end{aligned}
\end{equation}
where $L_g$ is the loss function, $G(s,a;\phi)$ represents the Guider's estimation for the true reward $r$ of state-action pair $s$-$a$ in data batch $\mathcal{D}$, parameters $\phi$ denotes the weights used by the Guider network and is updated via gradient descent method during training: 
\begin{equation} \label{eq:Reward Model update}
\phi = \phi  - \alpha_g {\nabla_\phi } L_g,
\end{equation}
 where $\alpha_g$ is the learning rate and ${\nabla_\phi} L_g$ is the gradient of the loss function $L_g$ with respect to parameters $\phi$. 

 During the online fine-tuning stage, we use the reward estimation provided by the Guider to guide agent action choices in exploration. Recall that we employ a $\epsilon$-greedy strategy to determine an agent's action for each edge when constructing the bipartite graph for order matching. Consider an agent $n$ in state $s_{n,t}$ at time $t$. When determining the agent's action for serving a rider $m$, we define a reward threshold $\widehat{r}$ and derive a filtered set of actions for whom the estimated rewards $G(s_t,a_{n,t};\phi)$ are larger than $\widehat{r}$: 
 \begin{equation}\label{filter_action}
A(s_{n,t}) = \{ a \mid G(s_{n,t},a_{n,t}) > \widehat{r} \}.
\end{equation}
 For exploitation (with probability $1-\epsilon$), the platform adheres to what it has learned from offline training and previous online tuning and the agent takes the action maximizing the expected gains $Q_{\pi}(s_{n,t},a_{n,t})$. In contrast, an agent in exploration (with probability $\epsilon$) would be assigned a random action in set $A(s_{n,t})$, instead of an arbitrary action described in Equation~\eqref{eq:Matching Value Function}.  Correspondingly, only the action whose estimated reward is larger than $\widehat{r}$ is assigned to a large upper bound Q-value.  This modifies the previously blindly optimistic exploration strategy to a more wisely optimistic exploration strategy. By introducing the "Guider" module, the weight associated with an edge connecting ride-pooling vehicle $m$ and rider $n$ is rewritten as follows: 
\begin{equation} \label{eq:reward guided exploration}
w(n, m) = \begin{cases} 
\underset{a_{n,t}}{\max} Q_{\pi}(s_{n,t}, a_{n,t}), & \text{with probability } 1 - \epsilon, \\
\overline{Q}, & \forall a_{n,t} \in A(s_{n,t}), \text{with probability } \epsilon.
\end{cases}
\end{equation}

As indicated by \eqref{filter_action}, the "Guider" module is mainly used to estimate the instant reward $r(s_{n,t},a_{n,t})$ of an agent for taking an action $a_{n,t}$ at the online stage. Although $r(s_{n,t},a_{n,t})$ can be accurately computed for each state-action pair (see Equation~\eqref{eq_reward}), computing them for thousands of agents and hundreds of riders online can be time-consuming due to the innate complexity incurred by vehicle routing optimization and transit network search. Applying function approximation $G(s_{n,t},a_{n,t})$ for reward removes such daunting tasks to the offline stage. Moreover, since the guide network training is a supervised learning task and is independent of the coordination policy, the reward guide network can be easily trained to a high level of accuracy and can extract significant intuition about the complex environment. As a result, agents can explore the system effectively under the guidance of the Guider during the online fine-tuning stage.

\begin{remark}
{We clarify that although the reward guider proposed in this paper draws inspiration from the existing literature on rewarding mechanisms for ride-sharing systems \cite{wang2018stable,agatz2011dynamic,hsieh2023improving,hsieh2024comparison}, our approach is conceptually distinct from those previously discussed. Existing studies focus primarily on the distribution of costs among drivers and passengers through methods like the Global Proportional (GP) Method \cite{hsieh2023improving} and the Driver Group-Passenger Group Proportional (DGPGP) Method \cite{hsieh2024comparison}, which aim to enhance user participation and the overall efficacy of ride-sharing services. In contrast, the reward guider introduced in our study is specifically designed to guide the exploration process within a reinforcement learning framework, thereby enhancing algorithmic efficiency. This Guider, which is developed and refined using historical vehicle trajectory data, employs a dynamically adjustable parameter, $\widehat{r}$, to steer vehicle agents toward actions that potentially serve both the platform's and passengers' interests during the online fine-tuning phase. Such an application of rewarding mechanisms to direct algorithmic exploration has not been previously explored, marking a novel contribution to the field.} 
\end{remark}

The remaining RL algorithms for fine-tuning are similar to those used in the offline stage. Notably, to encourage exploration, we adopt a decayed exploration rate at the early stage of online training: 
\begin{equation} \label{eq:epsilon_decay}
\epsilon = \max(\epsilon \cdot \beta, \epsilon_T)
\end{equation}
where $\epsilon$ is the current exploration rate, $\beta$ is the decay rate, and $\epsilon_T$ is a small predefined threshold exploration rate. Also, we employ experience replay to break the correlation of sequential experiences. This prevents the update process from becoming cyclical and counterproductive, ensuring a more stable and effective learning progression. 



\subsection{Overview of RG-CQL Framework}\label{subsec_overview}

The overview of our RG-CQL framework is depicted in Algorithm \ref{al: CPWT_RGCQL} {and Figure~\ref{fig:Overview of RG-CQL Framework}}, which embeds the key innovative concepts of our RG-CQL method delineated in Section \ref{subsec_offline} and Section \ref{subsec_online}.

In Algorithm \ref{al: CPWT_RGCQL}, Step 2 to Step 10 are dedicated to the offline training phase leveraging existing on-demand ride-hailing trip data, {corresponding to the RG-CQL training module in Figure~\ref{fig:Overview of RG-CQL Framework}}. Both the reward "Guider" and the CDDQN value function are trained based on a batch of state trajectories retrieved from existing datasets. Specifically, at each training step $t\in \{0,\Delta t,\cdots, T\}$, Step 6 samples a batch of trajectories regarding state transitions. Step 7 and Step 8 update the network parameters of CDDQN based on sampled trajectories, progressively identifying the best policy under conservative regularization. Step 9 updates the network parameters $\phi$ of the Guider to enhance the accuracy of the reward function regression via supervised learning.

\begin{figure}
  \centering
  \includegraphics[width=1\linewidth]{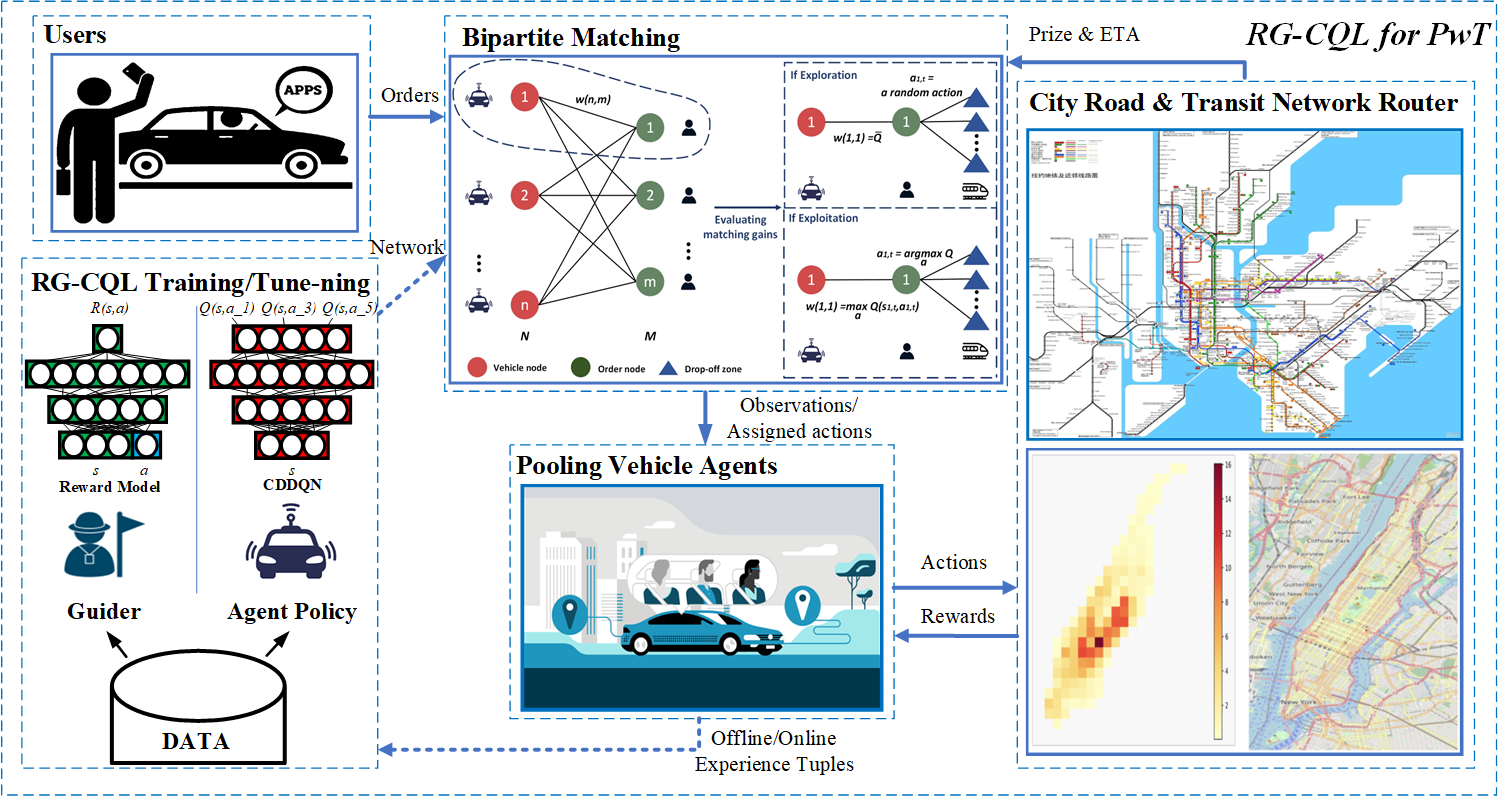}
      \caption{{Overview of RG-CQL Framework}}
  \label{fig:Overview of RG-CQL Framework}
\end{figure}



Step 11 to Step 28 transfer the training framework from the offline stage to the online stage for policy fine-tuning. Actions are assigned to agents based on the reward estimation from the Guider and the bipartite matching solutions. As new data transitions are accumulated, agents fine-tune their policies less conservatively, and the Guider could also concurrently update its understanding of the environment. To facilitate action evaluation, Step 12 introduces two routers: a vehicle router and a transit simulator. The vehicle router calculates optimal routes for ride-pooling vehicles, while the transit simulator estimates a rider's remaining transit time and the shortest path through the transit network (refer to Section \ref{subsec_simulationsetup} for details). Step 14 to Step 28 execute the online training and explore the best policy. In detail, each training episode commences with the updated exploration rate as specified in Equation~\eqref{eq:epsilon_decay} (Step 15). The bipartite graph is then constructed, {demonstrated in the bipartite match module in Figure~\ref{fig:Overview of RG-CQL Framework}}, with each edge representing a matching pair of agents and riders. The central platform assigns orders to vehicle agents using the ILP model outlined in Equations~\eqref{eq:ILP}, with reward estimation from Guider guiding agent exploration (see \eqref{eq:reward guided exploration}). In step 19, vehicle agents execute the assigned orders following the route computed by the vehicle router, {shown in the City Road and Transit Network Router in Figure~\ref{fig:Overview of RG-CQL Framework}}. Meanwhile, new experience tuples are collected and stored in the memory. With the newly accumulated experience, Steps 21 to 24 mirror the offline training process where both the "Guider" and the CDDQN value function are refined using batches of state trajectories drawn from the aggregated dataset. 

\begin{algorithm}[!htbp]
\caption{PwT\_RG-CQL Framework} \label{al: CPWT_RGCQL}
\begin{algorithmic}[1] 
\STATE Neural Network Initialization: CDDQN Training Net Parameter $\theta$, CDDQN Target Net Parameter $\theta^-$, Guider Net Parameters $\phi$.

\vspace{5mm} 
\STATE {\textbf{Offline Training Stage:}}
\STATE Dataset Initialization: past data transitions $D$ and sample Size $M$
\STATE Training hyper-parameter initialization: Conservative coefficient $C$, CDDQN update rate $\alpha_c$ and $\rho$, guider learning rate $\alpha_g$, number of training steps $T$
\FOR{$t = 0$ to $T$} 
    \STATE Sample $M$ experience tuples $(s, a, r, s')$ in $D$.
    \STATE Use Equation~(\ref{eq:CDDQN_Loss_Calculation}) to calculate $L_c$ and Equation~(\ref{eq:CDDQN update}) to update $\theta$.
    \STATE Update target network parameters $\theta^-$ using Equation~(\ref{eq:Polyak Average}).
    \STATE Use Equation~(\ref{eq:Reward Model Loss Calculation}) to calculate $L_g$ and Equation~(\ref{eq:Reward Model update}) to update $\phi$
\ENDFOR
\vspace{5mm} 

\STATE {\textbf{Online Fine-tuning Stage:}}
\STATE Initialization: Episode order requirements, vehicle router model, transit simulator, matching distance $R_{match}$, number of ride-pooling vehicles $N$. 
\STATE Hyper-parameters Initialization: Conservative Term $C$ as much smaller value, CDDQN Fine-tune Rate $\alpha$ and $\rho$, Guider Network Fine-tune Rate $\alpha_g$, Online Phase Exploration Rate $\epsilon$, $\epsilon_T$ with Exponential Decay Rate $\beta$, Memory Capacity $D$ and Memory Sample Size $M$.
\FOR{$e = 1$ to Episodes}
    \STATE Perform Exploration Decay via Equation~(\ref{eq:epsilon_decay}).
    \FOR{$t = 0$ to $t_{\text{terminal}}$ by $\Delta t$}
        \STATE Central platform updates order information, each vehicle’s location, and on-board passenger situations.
        \STATE Central platform assigns orders to vehicle agents according to ILP formulation in Equation (\ref{eq:ILP}) and (\ref{eq:reward guided exploration}) with the value estimation of the training network and guidance from Guider.
        \STATE Vehicles observe their orders and perform the assigned actions in the simulation platform and add every agent’s new experience tuple $(s, a, r, s')$ into the memory.
        \IF{memory size larger than $D$}
            \STATE Sample $M$ experience tuples $(s, a, r, s')$ in as mini-batch 
            \STATE Adopt Equation~(\ref{eq:CDDQN_Loss_Calculation}) and Equation~(\ref{eq:CDDQN update}) to update $\theta$.
            \STATE Update target network parameters $\theta^-$ using Equation~(\ref{eq:Polyak Average}).
            \STATE Use Equation~(\ref{eq:Reward Model Loss Calculation}) to calculate $L_g$ and Equation~(\ref{eq:Reward Model update}) to update $\phi$ if needed. 
        \ENDIF
        \STATE Based on the chosen action, central platform calculates the new route and estimated time of pick-up, drop-off, and transit.
    \ENDFOR
\ENDFOR
\end{algorithmic}
\end{algorithm}

\section{Simulation Experiments and Discussions}\label{sec_simulation}

In this section, we evaluate our proposed approach using real-world data across various scenarios. We begin by detailing the dataset and simulation parameters in Section 6.1. Next, in Section 6.2, we assess the performance of our approach with RL algorithms used in prior ride-hailing studies. {Furthermore, in Section 6.3, we evaluate the value of each component in our RG-CQL framework via ablations studies. In Section 6.4, we test the robustness of RG-CQL under variations of key heyper-parameters. Lastly, in Section 6.5, through comparing with State-of-the-art (SOTA) offline to online RL methods, we demonstrate the effectiveness of our approach in addressing offline to online gap in the context of ride-pooling with public transit.} Our simulation experiments serve two purposes. First, we demonstrate that integrating ride-pooling with transit services yields superior outcomes compared to systems offering pooling-only or non-pooling services. Second, we show that the proposed RG-CQL framework surpasses commonly employed RL algorithms (Online RL, Offline RL, and Offline to Online RL) in terms of training efficiency and effectiveness.



\subsection{Dataset and Simulation Setup}\label{subsec_simulationsetup}

\textbf{(a) Order data and study area} The simualtor is built based on trip request data extracted from the dataset presented in \cite{oda2018movi,al2019deeppool,haliem2021distributed}, which is sourced from the taxi trips of New York City~\cite{NYCTaxiData2018}. This dataset includes detailed trip-specific information such as pick-up and drop-off times, origin and destination geo-coordinates, trip distance, and duration. From this dataset, we extract data for trips occurring during the morning peak hour (8:00 AM - 9:00 AM) on May 4, 2016, with an average order density of around 271 trips per minute. For online fine-tuning, each training episode involves a sample of 95\% of these trips, totaling approximately 15,300 orders. The study area is Central Manhattan, which is partitioned into smaller grid zones with a resolution of 800m x 800m. The average number of trip requests originating from each zone per minute is calculated based on the extracted data. In the simulation study,  fifty-seven zones are selected for simulation. Fig.~\ref{fig:Demand Visualization} visualizes the order density of each demand zone using a heatmap, where the sidebar denotes the average number of orders received per minute. 

\begin{figure}[!htbp]
    \centering
    \subfigure[Order density]{
        \includegraphics[width=0.4\textwidth, height=9.1cm]{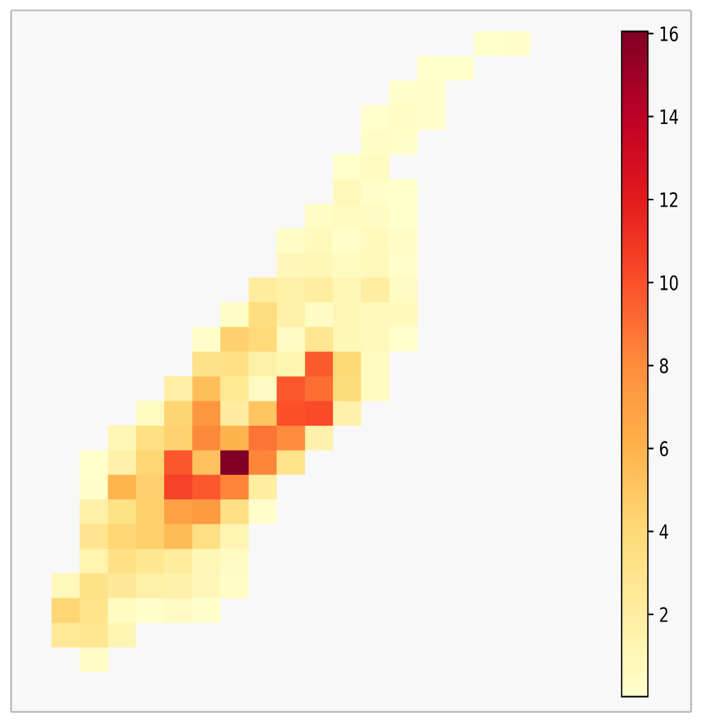}
        \label{fig:demand-visualization}
    }
    \subfigure[Central Manhattan]{
        \includegraphics[width=0.4\textwidth, height=9cm]{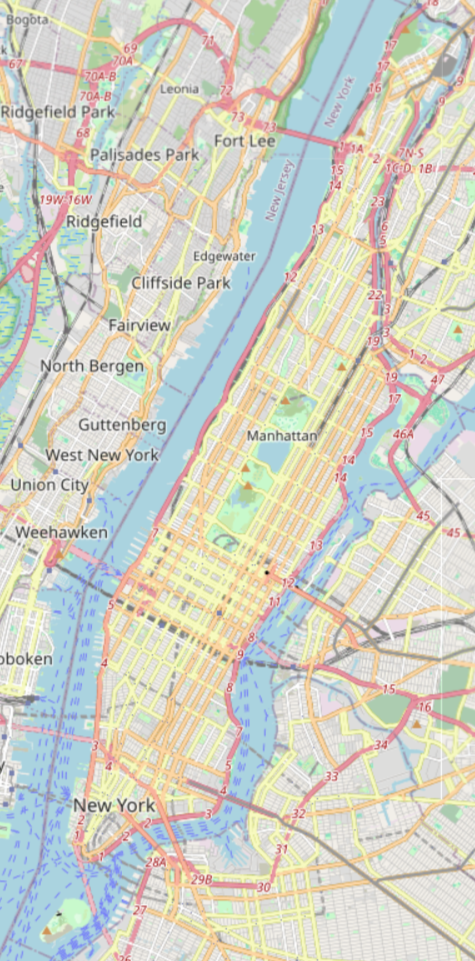}
        \label{fig:manhattan-map}
    }
    \caption{Study area and order density (Map from \cite{OpenStreetMap})}
    \label{fig:Demand Visualization}
\end{figure}


During simulation, we treat each trip request as an order from a rider, with the pick-up time serving as the trip request time. Based on the zone partitioning, the origin and destination zones for each trip are determined. Table~\ref{tab: Trip Example} presents a simulated order sample with all necessary information.

\textbf{(b) Road network and transit schedules}\quad Our routing optimization and vehicle navigation are based on the road network of Manhattan, which is obtained from OpenStreetMap. The information on transit services is obtained from the open-source project in~\cite{NYCTravelTimeMap} based on MTA schedules~\cite{MTASchedules} ,  which include the network of operated subway lines as well as the timetable for each line. Based on this information, we extract transit lines that provide trip services during the study period. In total, there are 29 subway lines in the entire New York City, encompassing over 380 unique subway stations. Distribution of transit stations in Central Manhattan is shown in Fig.~\ref{fig:Distribution of Transit Stations in Central Mahanttan}. 

\textbf{(c) Fleet size and matching operation} \quad To ensure that at least 60\% of riders can be served, we set the number of ride-pooling vehicles as 600 and the seat capacity of each vehicle as 3. The length of interval $\Delta t$ for order matching is 1 minute. The maximum matching distance $R_{match}$ is set to be 1.2km. During the simulation, we assume riders have a five-minute tolerance for matching time. Unmatched riders will exit the system if their time waiting to be matched exceeds this threshold. Regarding reward, the flag fare ${\beta _0}$ is set as 100, and the distance fare ${\beta _1}$is set as 40/km,  cost per unit waiting time ${\beta _2}$ is 5/min. We set the delay tolerance  $\kappa = 15$min. To encourage agents to serve more riders, we only penalize trip delay if the delay is higher than $\kappa$, and penalty coefficients ${\beta _3}$ and ${\beta _4}$ are set as 0/min and 10/min, respectively. {For solving the bipartite match, we utilized the Scipy library (version 1.13.1) which incorporates an optimizer named “$linear\_sum\_assignment$” to solve the ILP program~\cite{Scipy}. This optimizer leverages a modified Jonker-Volgenant algorithm~\cite{crouse2016implementing} to solve linear assignment problems as the ILP depicted in bipartite graphs. Notably, this algorithm has been proven to solve linear assignment problems within polynomial time. Our implementation shows this algorithm is highly efficient. During each decision-making round involving 600 vehicle agents and 500 orders, the optimizer required only about 1 second to solve the ILP. This duration includes the time taken for Q-value network and Guider network encoding for each potential vehicle-order link, which underscores the optimizer's computational efficiency in handling large-scale assignment problems.}


 \begin{table}[htbp]
\centering
\caption{An order sample in simulation}
\label{tab: Trip Example}
\small 
\begin{tabular}{@{}ccccccccccccccccc@{}}
\toprule
Request time & Origin Lat & Origin Lon & Destination Lat & Destination Lon & Origin Zone & Destination Zone \\
\midrule
8:10am & 40.727005 & -74.00322 & 40.731125 & -73.992233 & 7 & 13\\
\bottomrule
\end{tabular}
\end{table}

\textbf{(d) RL training } Our CDDQN utilizes a Multi-layer Perceptron (MLP) with a six-layer configuration. All layers except the final layer use the ReLU activation function to form the backbone of the neural network. The output of the MLP represents the Q-values for agent actions with the dimension being 58-by-1 (57 central demand zones for drop-off, one action for door-to-door services). Given the vehicle capacity of three, the input data for the CDDQN is a 1-by-14 tensor, which captures the state of each ride-pooling vehicle. For the Guider network, we utilized a similar six-layer MLP configuration. Its inputs include a 1-by-14 tensor representing the vehicle's state and a 1-by-1 tensor corresponding to a potential action, making the output an estimator for reward associated with a specific state-action pair. Regarding other hyperparameters, we set the sample size $M$ to be 1024. The learning rates $\alpha_c$ and $\alpha_g$ for CDDQN and Guider networks are set as 0.002 and 0.005, respectively. We employ a Polyak Averaging constant $\rho$ of 0.005 when updating the target network. During training, we use Mean Squared Error (MSE) for TD loss calculations and select Adam as the optimizer. The simulation computations are performed on an Intel Core i9-14900KF CPU, while all training processes are conducted on a NVIDIA GeForce RTX 4080 GPU. 

\textbf{(e) Routing and vehicle navigation} \quad Our vehicle router model implementes the Open Source Routing Machine (OSRM) model~\cite{ProjectOSRM} via Docker to provide real-time ride-pooling vehicle route guidance on the real city road network and to estimate passengers' onboard time. OSRM processes raw OpenStreetMap data~\cite{OpenStreetMap} to extract city road network, representing the network as a directed graph where intersections serve as nodes and road segments serve as edges. Each edge is enriched with metadata including road distance, speed limits, and vehicle travel time. When handling TSP routing requests from ride-pooling vehicles, OSRM employs different strategies based on the number of waypoints. When the number of waypoints is larger than 10, it utilizes a greedy heuristic approach, specifically the farthest-insertion algorithm to compute the shortest route.  For fewer than 10 waypoints, OSRM applies a brute force method to find the optimal solution~\cite{OSRMAPI}. It's important to note that as solving TSP is NP-hard, OSRM provides an approximation rather than an guaranteed optimal solution for larger problems. This approach strikes a balance between computational efficiency and route optimization, enabling the system to handle real-time requests effectively while still producing high-quality route recommendations. An example for routing a ride-pooling vehicle from two riders' common origins to their drop-off locations are shown in Table~\ref{tab:OSRM Request}\footnote{The Url request can be injected to the local OSRM API \cite{ProjectOSRM,OSRMAPI}. The API would returns a JSON object containing detailed information about the route.} 
By leveraging OSRM's capabilities, we can generate realistic travel time estimates and route guidance, enhancing the accuracy of our city network simulation. {We evaluated the OSRM API against various heuristic baselines \cite{Openrouteservice} and observed a relative error rate of less than 3\%, confirming its high accuracy.}

\begin{table}[h]
    \centering
    \caption{Example OSRM routing request}
    \label{tab:OSRM Request}
    \begin{tabular}{
      >{\raggedright\arraybackslash}p{4cm} 
      >{\raggedright\arraybackslash}p{4cm} 
      >{\raggedright\arraybackslash}p{8.5cm} 
    }
        \toprule
        {\small Coordinates of origin} & {\small Coordinates of two drop-off points} & {\small Url request} \\
        \midrule
        \small {(40.735212, -73.995230)} & \small{(40.75270, -73.986064), (40.790758, -73.951609)} & \small{"http://localhost:5000/trip/v1/driving/-73.995230, 40.735212; -73.986064, 40.75270; -73.951609, 40.790758 ?roundtrip=false\&source=first\&annotations=true"} \\
        \bottomrule
    \end{tabular}
\end{table}

For transit simulator, we develop a Transit ETA model to estimate a rider's trip time on transit and to compute her shortest path over transit networks. The Transit ETA model is developed based on NYC's transit schedule~\cite{MTASchedules} and open-source project~\cite{NYCTravelTimeMap} designed for computing travel times between two subway stations.  As the first step, we construct a transit graph with nodes representing transit stations and edges representing various types of connections. Two nodes are created for each station with one node "station\_id+LINE" representing one transit line direction and "station\_id-LINE" representing the opposite direction. The graph incorporates four exclusive sets of edges, and each edge is associate with a value denoting travel time on the edge:
\begin{itemize}
    \item Boarding edges: Connect stations to trains (station\_id → station\_id\textpm LINE), weighted with a fixed frequency value of 160 seconds.
    \item Exiting edges: Link trains back to stations (station\_id\textpm LINE → station\_id), weighted as 40 seconds to represent passenger exiting cost.
    \item Travel edges: Connect stations along the same line (station\_id1\textpm LINE → station\_id2\textpm LINE), weighted with the calculated past trip travel time near 9:00 AM between consecutive stops.
    \item Transfer edges: Edges to facilitate changes between lines at stations, weighted with the minimum transfer time derived from transfer information.
\end{itemize}

 
Based on the constructed graph, Dijkstra's algorithm is used to find the fastest path between stations for each transit routing request. The transit router enhances the routing process by incorporating walking distances to and from stations, ensuring the calculation of the most efficient overall route. This approach combines public transit use with pedestrian segments (assuming a pedestrian walking speed of 3.6 km/h) to provide comprehensive door-to-door journey planning. By comparing our subway route and ETA time with that of google map~\cite{GoogleMaps} in Table~\ref{tab:time_comparisons}, we demonstrate the accuracy of our transit model. 
\begin{table}[h]
    \centering
    \caption{Transit time using ETA model and Google map}
    \label{tab:time_comparisons}
    \resizebox{0.8\textwidth}{!}{%
        \begin{tabular}{ccc}\toprule
            {Example Query} & {Transit ETA Model} & {Google Map} \\ \midrule
            From the Met Cloisters (40.865491, -73.927271) \\ to Holcombe Rucker Park (40.830915, -73.936589) & 24 minutes & 25 minutes \\ \bottomrule
        \end{tabular}
    }
\end{table}

Fig.~\ref{fig:Operation Visualization} provides a example for vehicle route navigation. As depicted in the figure, a ride-pooling vehicle is designated to serve two riders whose destinations are represented by purple icons. The vehicle agent adheres to the route generated by our OSRM router (dipicted by the blue line) and drops off a passenger at the subway station marked by the red icon. The drop-off station is determined by our transit simulator based on the agent's action. After being dropped off by the ride-pooling vehicle, the passenger will use the transit services to reach her final destination, following the path (shown by the purple links) provided by the transit simulator. If a purple icon overlaps with a red icon, it indicates that the vehicle agent deliver the passenger directly to her final destination without involving transit services. 



\begin{figure}[!htbp]
    \centering
    \subfigure[Transit station distribution in central Manhattan]{
        \includegraphics[width=0.4\textwidth, height=8cm]{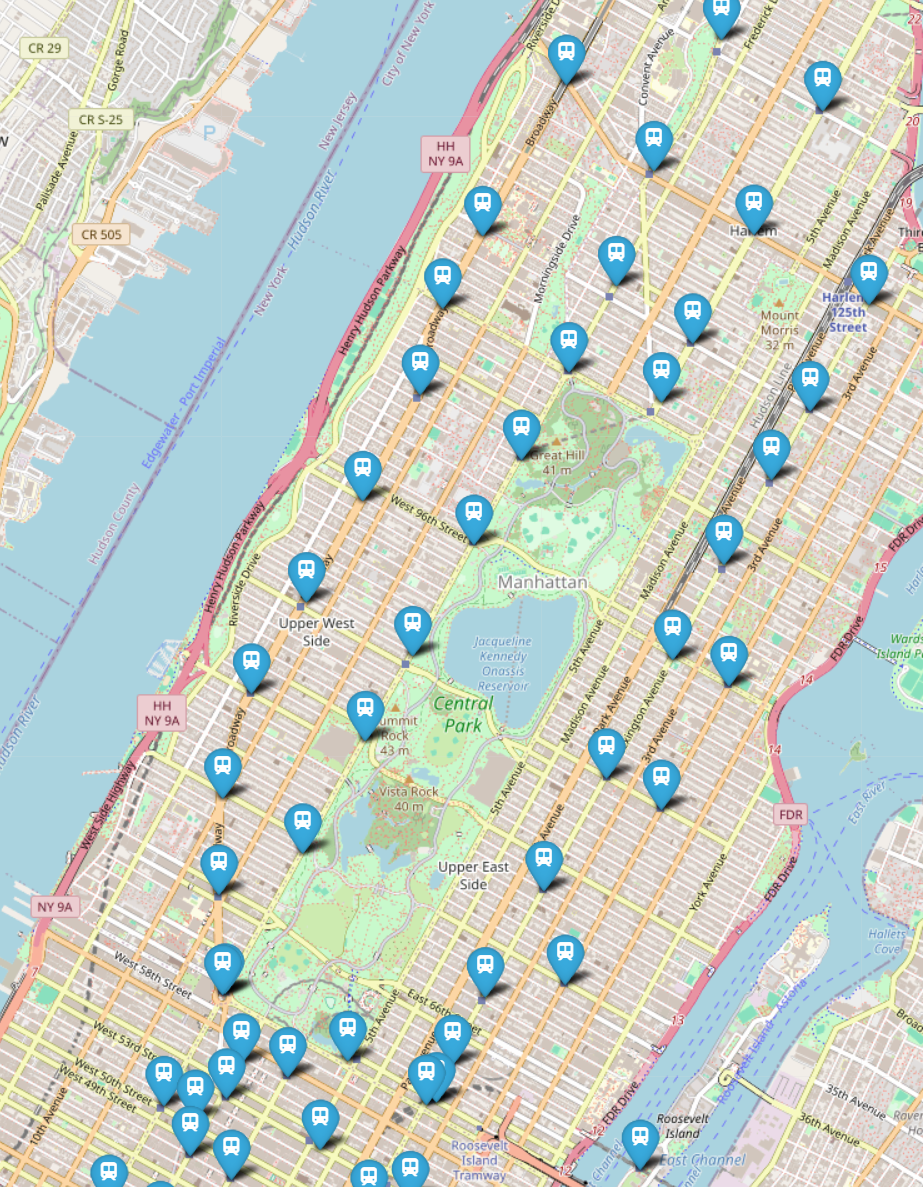}
        \label{fig:Distribution of Transit Stations in Central Mahanttan}
    }
    \subfigure[A routing example visualization]{
        \includegraphics[width=0.45\textwidth, height=8cm]{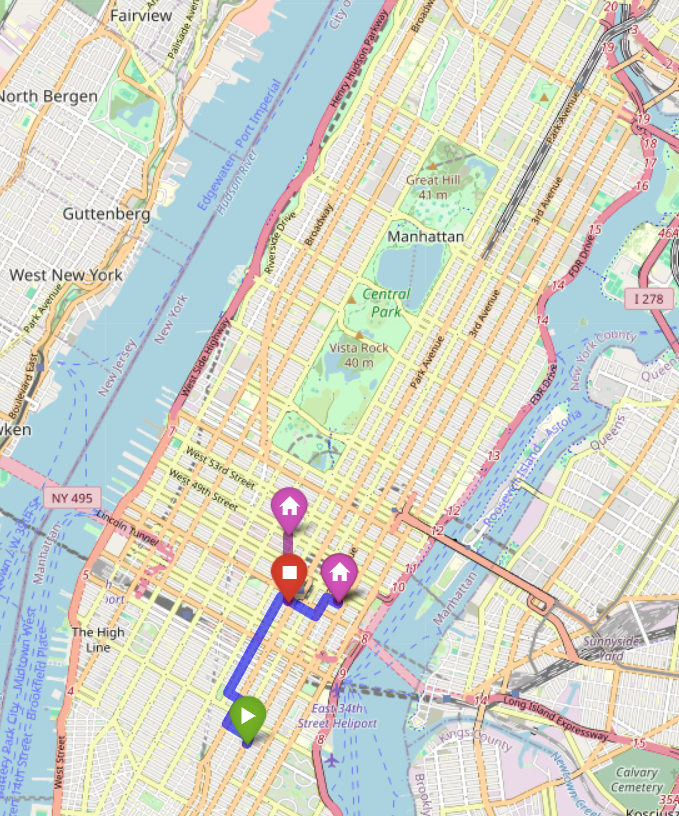}
        \label{fig:Operation Visualization}
    }
    \caption{Map of the transit stations and a visualization example for coordinated ride-pooling and transit operations}
    \label{fig: Simulator Operation Visualization}
\end{figure}


\subsection{Comparison of Different Ride-Hailing Methods in Multi-modal Transportation}\label{subsec_RLbaseline}
In this subsection, we compare our approach with cutting-edge baseline methods that vary in multi-modal transportation systems. The aim is to demonstrate the superiority of the proposed coordinated pooling-transit services and RG-CQL framework in enhancing ride-hailing system performance. To distinguish between our comparison cases, we use "A\_B" notation, where A represents the ride-hailing service mode ("PwT" for the proposed coordinated pooling-transit services, "P" for purely ride-pooling services, and "NPwT" for coordinated non-pooling and transit services ) and B represents the method (RG-CQL, "Online RL",  "Greedy", and  {"Insertion" method}). 

The following outlines the comparison cases and discusses how our framework adapts to each baseline method for evaluation:  
\begin{itemize}
    \item \textbf{{PwT\_Online RL}}: The service mode in this benchmark aligns with the description in Section 3. The MDP model and order matching model are consistent with those previously introduced. However, only the online RL algorithm DDQN is executed to learn the value function and optimal policy. 
    \item \textbf{{PwT\_Greedy} \cite{stiglic2018enhancing,gu2024algorithms}}: We train a reward model for the environment through online iteration using a neural network whose architecture is identical to our Guider. During exploitation, the agent selects actions that maximize the estimated reward. The platform optimizes total rewards in order matching without considering long-term gains. 
    \item \textbf{{{PwT\_Insertion}} \cite{simonetto2019real,edirimanna2024integrating}}: {We solve linear assignment problems to match drivers with riders on a one-to-one basis, employing the sequential graph approach from \cite{simonetto2019real}. We have adapted this method to our context by incorporating flexible drop-off locations through heuristic insertion, which is suitable for large-scale ride-pooling, as described in \cite{edirimanna2024integrating}. To remain consistent with other baselines, we impose strict time constraint as 15 minutes, only considering transit stations that meet customers' arrival deadlines within their trip windows.}
    \item \textbf{{P\_Online RL} \cite{al2019deeppool}}: Ride-pooling vehicles must deliver riders to their destinations, limiting an agent's action to door-to-door services. Online RL algorithm DDQN is executed to learn value function and optimal policy.  
    \item \textbf{{NPwT\_Online RL} \cite{feng2022coordinating}}: Pooling is prohibited so that a vehicle has to complete its current trip before becoming available for matching. Passengers can be dropped off at intermediate transit stations or their final destinations. Online RL algorithm DDQN is used for training. 
     \item \textbf{{PwT\_RG-CQL}}: We aggregate vehicle trajectories from the mixture of four policies above as synthetic data (details are introduced in next subsection \ref{subsec_Sensitivity}) and adopt our proposed approach outlined in Algorithm \ref{al: CPWT_RGCQL} to train the policy. 
\end{itemize}


Table~\ref{tab:algorithm_component} summarizes the key features of each method listed above. 
\begin{table}[ht]
\centering
\caption{Key components of different methods}
\label{tab:algorithm_component}
\begin{tabular}{@{}lccccc@{}}
\toprule
                        & Pooling & With Transit & Myopic & Offline Learning & Online Learning \\ 
\midrule
PwT\_Online RL & $\checkmark$ & $\checkmark$ & $\times$ & $\times$ & $\checkmark$ \\
PwT\_Greedy    & $\checkmark$ & $\checkmark$     & $\checkmark$ & $\times$ & $\checkmark$ \\
{PwT\_Insertion} & $\checkmark$ & $\checkmark$  & $\checkmark$ & $\times$ & $\times$ \\
P\_Online RL   & $\checkmark$     & $\times$     & $\times$ & $\times$ & $\checkmark$ \\
NPwT\_Online RL& $\times$     & $\checkmark$ & $\times$ & $\times$ & $\checkmark$ \\
PwT\_RG-CQL    & $\checkmark$ & $\checkmark$ & $\times$ & $\checkmark$ & $\checkmark$     \\
\bottomrule
\end{tabular}
\end{table}

The networks are trained over one thousand episodes for the online RL approach. We initialize the exploration rate at 1 and gradually decrease it to 0.005, employing a decay rate of 0.995 to entitle an abundant exploration budget for fair comparisons. Fig.~\ref{fig:Training of Different Ride-Sharing Algorithms} shows the variations in the accumulative total reward with training progression, where the shaded area represents the range between the maximum and minimum values observed across the nearest 25 episodes. As shown in Fig.~\ref{fig:Training of Different Ride-Sharing Algorithms}, the proposed RG-CQL demonstrates superior learning efficiency and effectiveness compared with all other baseline methods.\footnote{Note that the starting point of the training curve for the proposed RG-CQL method is better than other benchmark methods because we combine offline training with online fine-tuning, where the offline training phase offers a good starting point for the online fine-tuning phase. }


\begin{figure}[h]
  \centering
  \includegraphics[width=0.6\linewidth]{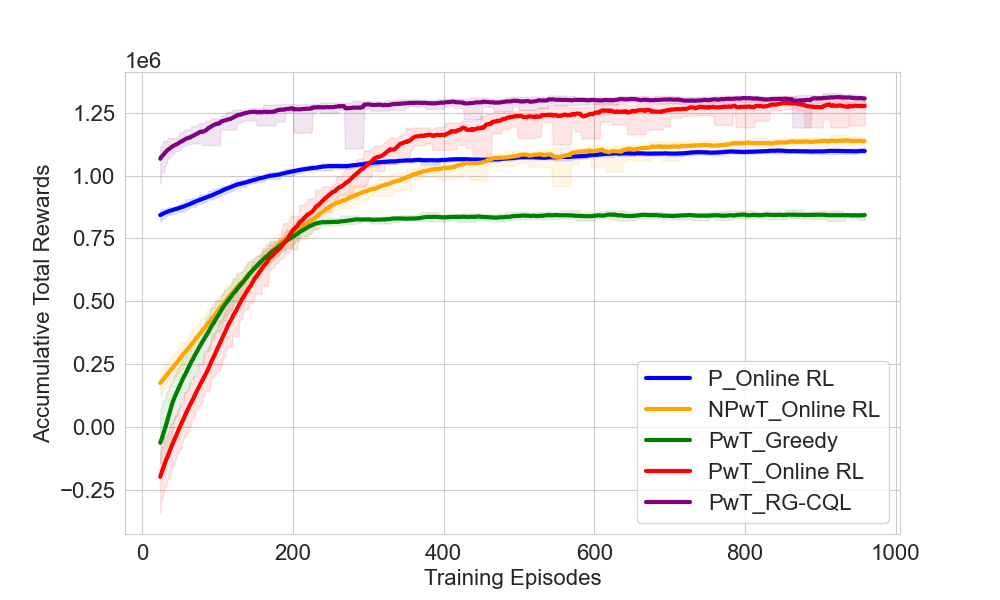}
  \caption{Training comparison of different Ride-hailing modes and RL methods}
  \label{fig:Training of Different Ride-Sharing Algorithms}
\end{figure}

To demonstrate the efficacy of our RG-CQL framework in enhancing ride-hailing system performance, we employ a range of metrics, including (i) Order Service Rate, which is calculated as the ratio of served orders regarding the total order population; (ii) Overestimation Rate, refers to the discrepancy between the RL agent's estimated accumulative total rewards and the true accumulative total rewards; (iii) Accumulative Total Rewards; (iv) Average Detour, refers to riders' average additional travel time in minutes when they are served by a specific ride-hailing mode as compared to the non-pooling door-to-door  service mode. 

Table \ref{tab:Comparison of Different Ride-Sharing Modes} compares ride-hailing system performances across different comparison cases. One observation is that coordinating transit with ride-hailing service, either pooling or non-pooling, yields a higher order service rate and platform profits than purely ride-pooling services. Compared to purely ride-pooling services (P\_Online RL), although riders experience higher detour time, coordinating non-pooling services with transit can increase the order service rate by 10\%. This is further increased to approximately 31\% when ride-pooling is permitted. Such benefits can be explained by the increased flexibility in riders' drop-off locations when ride-hailing is coordinated with transit. Compared to purely door-to-door ride-pooling services, more vehicles would become available for matching when riders can be dropped off before reaching their destinations.

Another observation is that RG-CQL method demonstrates significant improvements regarding total rewards and overestimation rate compared to all other cases using online RL DDQN algorithms. More specifically,
the proposed PwT\_RG-CQL method yields the highest total accumulative rewards among all comparison cases. For the coordinated service mode, the reward under PwT\_RG-CQL mode surpasses those under NPwT\_Online RL and PwT\_Online RL by 17\% and 22\%, respectively. In addition, the overestimation rate under PwT\_RG-CQL is much lower than those under NPwT\_Online RL and PwT\_Online RL, thanks to our introduction of conservative regularization. Last but not least, our RG-CQL manages to push average detour closest to that of the passenger tolerance. {Notably, by fully encoding future uncertainties, PwT\_RG-CQL significantly outperforms the optimization baseline PwT\_Insertion method in crucial operational metrics such as service rate and total rewards. This further confirms our method efficacy in managing large-scale ride-hailing services, demonstrating its superior strategic deployment and optimization capabilities.}



\begin{table}[htbp]
\centering
\caption{Ride-hailing system performance under RG-CQL and baseline methods}
\label{tab:Comparison of Different Ride-Sharing Modes}
\begin{tabular}{@{}lcccc@{}}
\toprule
      Method & Service Rate & Overestimation Rate & Accumulative Total Rewards & Avg. Detour \\\midrule
      P\_Online RL & 61.1\% & \textbf{2.3\%} & 1,095,135 & \textbf{3.62} \\
      NPwT\_Online RL & 71.0\% & 6.2\% & 1,143,557 & 15.41 \\
      {PwT\_Insertion} & 61.1\% & NA & 883,440 & 10.00 \\
      PwT\_Online RL & 92.0\% & 9.1\% & 1,281,804 & 16.00 \\
      PwT\_RG-CQL & \textbf{92.2\%} & 3.5\% & \textbf{1,336,361} & 15.37 \\\bottomrule
\end{tabular}
\end{table}


\subsection{{Ablation Study of the Proposed RG-CQL Framework}}\label{subsec_ablation}

This subsection aims to demonstrate the efficiency and robustness of our RG-CQL framework {via ablation studies, especially the effectiveness of incorporating the Guider, offline training initialization and online training, respectively, into the RL process. For the studies,} based on the training results of the four baseline algorithms (Section \ref{subsec_RLbaseline}), we construct two synthetic datasets T1 and T2 for offline training:  
\begin{itemize}
    \item \textbf{T1: Expert but limited coverage dataset}. This dataset aggregates 100,000 data transitions, 90\% of which are derived from the baseline algorithm PwT\_Online RL, and other 10\% from the random exploration policy where agents randomly choose actions. Although most data samples are from an expert-driven policy, the dataset only offers a limited coverage of simulation transitions.
    \item \textbf{T2: Large coverage but chaotic dataset}. The dataset comprises 200,000 data transitions, with 25\% of data samples from each of the following policies: PwT\_Online RL, P\_Online RL, PwT\_Greedy, and the random exploration policy where agents randomly choose actions. While offering a more comprehensive representation of the simulated environment compared to T1, the data sources in this dataset are notably noisier.
\end{itemize}

{First, to showcase the effectiveness of incorporating the Guider, we train our RG-CQL algorithm and classic CQL~\cite{kumar2020conservative} algorithms\footnote{We also train off-policy algorithms such as DDQN and Expected SARSA on the same dataset; however, these methods suffered significantly from out-of-distribution (OOD) issues, leading to unstable performance.} that are fined tuned without using Guider. At the offline training stage, both algorithms are trained using dataset T1. For fine-tuning RG-CQL, we set the reward threshold $\widehat{r}$ to be 100. To make our results more representative, we fine-tune both the RG-CQL and naive CQL algorithms under two exploration rates: a high initial exploration rate of 10\% and a low initial exploration rate of 5\%.The training results are shown in Fig.~\ref{fig:Comparison of CQL and RG-CQL} and Table~\ref{tab:Comparison of Online Fine-tuning of RG-CQL and Baselines under Different Exploration Rate}. There, "highexp" and "lowexp" correspond to high and low initial exploration rates, respectively. As shown, RG-CQL framework outperforms CQL and PwT\_Online RL algorithms by delivering higher total accumulative rewards while significantly reducing online training time. Specifically, under both high and low initial exploration rates, RG-CQL achieves higher accumulative rewards compared to CQL. With a high initial exploration rate, RG-CQL boosts accumulative rewards by approximately 5.2\% compared to CQL, a figure that rises to about 6\% under a low initial exploration rate. In the absence of the Guider, CQL even falls short of PwT\_Online RL in terms of accumulative rewards. Moreover, our proposed RG-CQL framework significantly decreases online fine-tuning time compared to CQL, by over 40\% under low initial exploration rate and 61\% under high initial exploration rate. This indicates that including a Guider in offline RL fine-tuning effectively addresses slow learning and initial unlearning challenges. }

\begin{figure}[h]
  \centering
  \includegraphics[width=0.6\linewidth]{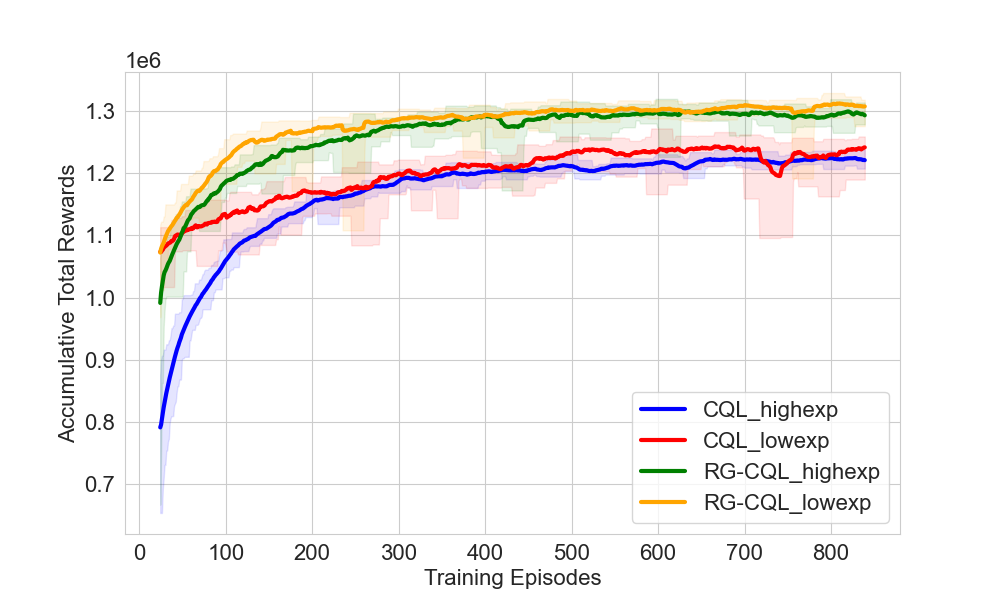}
  \caption{Fine-tuning of RG-CQL and CQL under different initial exploration rate}
  \label{fig:Comparison of CQL and RG-CQL}
\end{figure}

\begin{table}[ht]
\centering
\caption{Comparison of online fine-tuning of RG-CQL and baselines under different initial exploration rate}
\label{tab:Comparison of Online Fine-tuning of RG-CQL and Baselines under Different Exploration Rate}
\begin{tabular}{@{}lcc@{}}
\toprule
Method & Accumulative Total Rewards & Episode to Reach 1,200,000 Total Reward \\
\midrule
RG-CQL\_highexp & 1,294,622 (\textbf{+5.99\%}) & 121 (\textbf{-73.5\%}) \\
RG-CQL\_lowexp & 1,317,564 (\textbf{+7.87\%}) & 85 (\textbf{-81.3\%}) \\
CQL\_lowexp & 1,240,408 (\textbf{+1.55\%}) & 307 (\textbf{-32.7\%}) \\
CQL\_highexp & 1,221,457 (\textbf{0.00\%}) & 361 (\textbf{-20.8\%}) \\
PWT\_Online RL   & 1,275,987 (\textbf{+4.46\%}) & 456 (\textbf{00.0\%}) \\
\bottomrule
\end{tabular}
\end{table}

{Secondly, to highlight the efficiency of the RG-CQL solution due to the integration of offline training within the RL framework, we have included results from the PwT\_Online RL algorithm as a baseline, which lacks offline training.  The impact of incorporating offline training in RL is evident when comparing the online training time for each algorithm (see Table \ref{tab:Comparison of Online Fine-tuning of RG-CQL and Baselines under Different Exploration Rate}). Both RG-CQL and CQL exhibit reduced online fine-tuning time compared to PwT\_Online RL. Noteworthy is the substantial enhancement in online fine-tuning time achieved by RG-CQL, accelerating the online training time by 81.3\% over PwT\_Online RL when operating with a high initial exploration rate. }




{Furthermore, we examine two scenarios to evaluate the importance of online fine-tuning. Specifically, we separately train the algorithm using datasets T1 and T2 on the offline stage and subsequently fine-tune the algorithm online. During online training, we set the initial exploration rate to 10\% and maintain the reward threshold $\widehat{r}$ at 100. Fig.~\ref{fig:Comparison of RG-CQL and Baselines under Different Initialization} illustrates the changes in total cumulative rewards throughout the training process and Table~\ref{tab:Comparison of Online Fine-tuning of RG-CQL with Different Initialization} compares the cumulative reward at the start and end of the online training stage across different scenarios. As shown, the RG-CQL algorithm yields a consistently high level of total rewards after online fine-tuning compared with offline training initialization, regardless of the dataset used. Notably, the total cumulative reward after offline training amounts to approximately 89\% of that achieved after online training, due to the incompleteness of offline data batch. This validates the importance of online fine-tuning in RG-CQL. Moreover, despite the distinct inputs provided by two datasets after offline training, we could also observe the total cumulative rewards are nearly identical for RG-CQL\_T1 and RG-CQL\_T2 following online fine-tuning. This observation also confirms the robustness of our RG-CQL framework concerning variations in offline training datasets. }

\begin{figure}[h]
  \centering
  \includegraphics[width=0.6\linewidth]{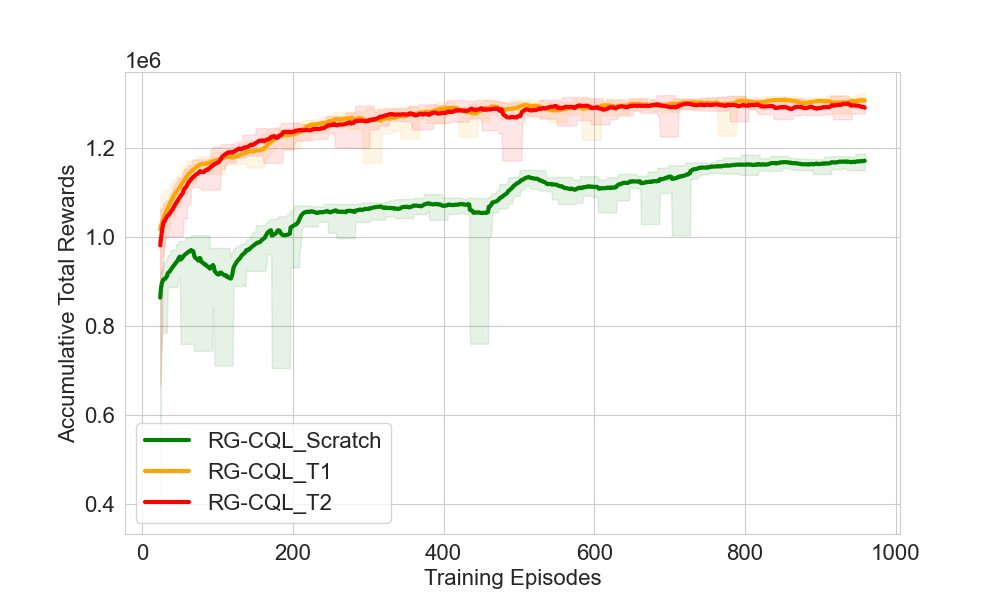}
  \caption{Comparison of RG-CQL and baselines under different initialization inputs}
  \label{fig:Comparison of RG-CQL and Baselines under Different Initialization}
\end{figure}

\begin{table}[ht]
\centering
\caption{Comparison of online fine-tuning of RG-CQL under different initialization inputs}
\label{tab:Comparison of Online Fine-tuning of RG-CQL with Different Initialization}
\begin{tabular}{@{}lc@{}}
\toprule
Method & Accumulative Total Rewards (Initial $--->$ Final) \\
\midrule
RG-CQL\_T1 & 1,161,029 $--->$ 1,296,078 (+10.7\%) \\
RG-CQL\_T2 & 1,064,764 $--->$ 1,294,622 (+10.6\%) \\
RG-CQL\_Scratch & 258,025 $--->$ 1,170,969 (0.00\%) \\
\bottomrule
\end{tabular}
\end{table}

{Lastly, to further emphasize the value of offline training, especially in scenarios where exploration budget might be limited, we introduce and add another scenario called "RG-CQL\_Scratch" where the algorithm is trained from scratch, shown in Fig.~\ref{fig:Comparison of RG-CQL and Baselines under Different Initialization} and Table~\ref{tab:Comparison of Online Fine-tuning of RG-CQL with Different Initialization} as well. No datasets are used for offline training under scenario "RG-CQL\_Scratch", requiring agents to learn the value function and policy entirely during the online training phase. This approach aligns with the method presented in~\cite{feng2022coordinating}, except that we consider and strengthen agent exploration (the exploration rate is set as 10\%). Compared to the "RG-CQL\_Scratch" scenario, the RG-CQL algorithm enhances the cumulative reward by more than 10\% and reduce fluctuations in total rewards during the online fine-tuning stage. This further suggests that the offline training does offer a good initial policy for online fine-tuning by learning from existing data.
}

\subsection{{Sensitivity Analysis of RG-CQL}}\label{subsec_Sensitivity}
{To assess the robustness of the RG-CQL algorithm, we conducted three sets of sensitivity analyses focusing on the impact of variations in key hyperparameters and real-world scenarios on the RG-CQL online fine-tuning process. These analyses explored the effects of changes in the learning rate (\(\alpha_c\)), experience memory capacity (\(D\)), and the percentage of orders that require pooling-only services (\(p_{pool}\)).}

{First, we evaluated the impact of different learning rate settings on the RG-CQL framework by comparing the original online learning rate of $\alpha_c = 0.002$ with $\alpha_c = 0.0015$ and $\alpha_c = 0.0025$. The training curves, illustrated in Figure~\ref{fig:Comparison of RG-CQL under Different Hyper}(a), indicate that the fine-tuning performance remains consistently stable across these variations. Additionally, we conducted a sensitivity analysis on experience memory capacity $D$ by retraining our model with capacities of $D = 8,000$ and $D = 12,000$, and compared these with the initial setting of $D = 10,000$. The results, depicted in Figure~\ref{fig:Comparison of RG-CQL under Different Hyper}(b), also demonstrate that training performance does not significantly vary with changes in experience memory capacity. The two results confirm the robustness of our RG-CQL framework under different learning hyper-parameters.}

\begin{figure}[ht]
  \centering
  \begin{minipage}{0.48\linewidth}
    \centering
    \includegraphics[width=\linewidth, height=0.7\linewidth]{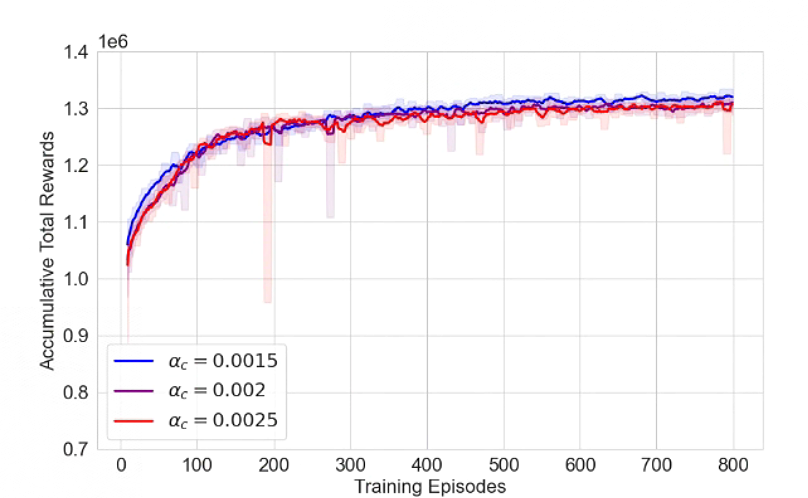}
    {\textbf{(a)} RG-CQL under different learning rates $\alpha_c$}
  \end{minipage}
  \hfill
  \begin{minipage}{0.48\linewidth}
    \centering
    \includegraphics[width=\linewidth, height=0.7\linewidth]{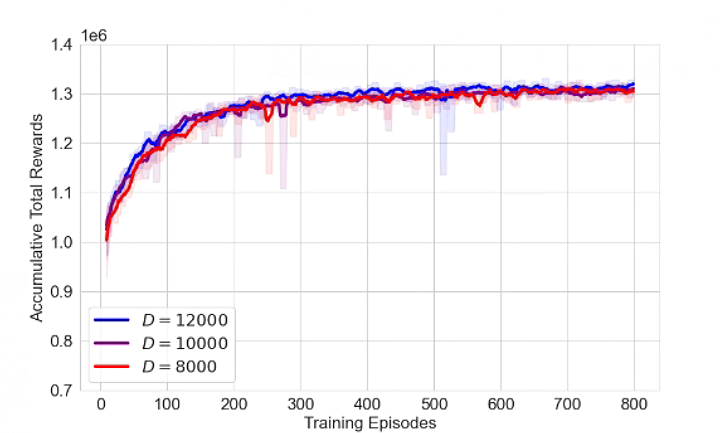}
    {\textbf{(b)} RG-CQL under different  capacities $D$}
  \end{minipage}
  \caption{{Sensitivity analyses of RG-CQL framework under various hyper-parameters (With rolling window set as 10 episodes)}}
   \label{fig:Comparison of RG-CQL under Different Hyper}
\end{figure}

{Furthermore, we conducted another sensitivity analysis to examine the impact of the variable \( p_{pool}\), which represents the percentage of orders that require pooling-only services. This analysis was prompted by the realization that in many real-world scenarios, the choice between participating in multimodal transportation or being driven directly to a destination is not typically decided by the driver or the system but rather by the passengers based on their personal preferences. Therefore, we introduced \( p_{pool}\) as a means to quantify the proportion of customers who prefer exclusive pooling services, while others might defer to the driver’s discretion. This adjustment enhances the realism of our simulation environment and better aligns it with potential real-world deployment scenarios. Specifically, we sampled and labeled different percentages of orders that require pooling-only service. For each labeled order, during the bipartite matching process, only the first action (\( a = 0 \)) and its corresponding q-value $Q(s,a=0)$ are considered valid. We retrained our RG-CQL framework in online fine-tuning stage under three variations of \( p_{pool}\): \( 10\%\), \( 20\% \), and \( 30\%\), and compared their final performance to our original setting where \( p_{pool}= 0 \). The results of this comparison are presented in Table \ref{tab:pooling_ratios}. As observed in our experimental results, despite the increasing complexity and less pooling chances imposed by the growth of \( p_{pool}\), our RG-CQL framework successfully adapts and identifies an optimal policy. This capability is evident from the strong performance metrics recorded across various scenarios: total rewards, order service rate, and average passenger detour. Notably, even as \( p_{pool} \) increases from 0\% to 30\%, the framework maintains commendable levels of service rate and minimizes detours effectively. These results collectively validate the robustness and reliability of our RG-CQL framework in handling dynamic and challenging real-world conditions.}

\begin{table}[h!]
  \centering
  \caption{{Impact of Different Percentage of Orders Requiring Pooling-Only Service on RG-CQL}}
  \label{tab:pooling_ratios}
  \begin{tabular}{lcccc}
    \toprule
    Percentage $p_{pool}$ & Total Rewards & Order Service Rate & Average Passenger Detour \\
    \midrule
    0\%  & 1,336,361 & 92.90\% & 15.37 min \\
    10\% & 1,318,879 & 89.80\% & 14.38 min \\
    20\% & 1,271,447 & 84.90\% & 13.44 min \\
    30\% & 1,259,594 & 82.60\% & 12.45 min \\
    \bottomrule
  \end{tabular}
\end{table}

\subsection{{Comparison of RG-CQL to other SOTA Offline to Online RL Baselines}}\label{subsec_O2O}
{To better support our analysis in Section~\ref{subsec_Offfline_RL_Review} and demonstrate the superiority of our approach against SOTA offline to online RL algorithms in large scale multimodal transportation systems, we trained two SOTA offline-online RL baselines, detailed as below:}
    \begin{itemize}
         \item {\textbf{{Cal-q}}: We train the CDDQN policy according to the proposed less conservative regularization in \cite{nakamoto2024cal}, where the q-value for unseen state-action pairs will be bounded with pre-trained reference policy value function in the offline stage. For fair comparison to our reward guider, the reference policy value function for Cal-q is trained under the greedy sample policy. During the online fine-tuning stage, we do not utilize Guider to guide agent's exploration.} 
        \item {\textbf{{Hybrid-q}}: As proposed in~\cite{song2022hybrid}, we train the DDQN policy directly in the online stage, but preload and keep offline data batch in the experience memory batch.}
    \end{itemize}
{The training plot comparison of our RG-CQL with Cal-q and Hybrid-q is shown in Figure \ref{fig:SOTA offline-online RL baselines}. We could observe from the plot that compared with SOTA offline-online RL algorithms such as Cal-q and Hybrid-q, the proposed method (i.e., RG-CQL) manages to effectively address the offline to online dilemma in the context of coordinating ride-pooling with public transit by having 6.7\% and 16.2\% improvement in total rewards respectively. Since Hybrid-q proposed by \cite{song2022hybrid} involves loading pre-collected data trajectories as an initial experience batch for off-policy RL methods and then conducting training as usual, though simple and efficient, its performance is still upper-bounded by SOTA pure offline RL pipelines. For Cal-q, because the lower-bound of the pre-trained reference value function is crucial for the proposed framework and how inaccurate reference value function will impact the performance of the policy remains unknown, this method requires expert prior knowledge of the sample policy or environment system. The challenges above limit the adoption of these popular offline to online RL methods in our setting, which is more complex due the the necessity of coordinating ride-pooling with public transit.}

\begin{figure}
  \centering
  \includegraphics[width=0.6\linewidth]{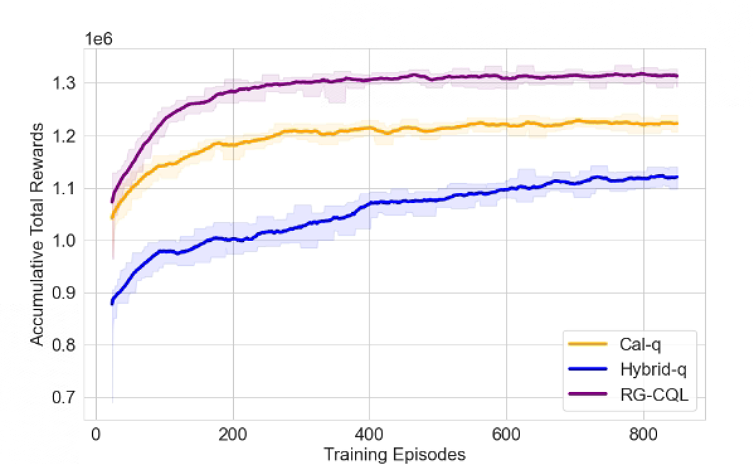}
  \caption{{Comparison of RG-CQL with other SOTA offline to online RL baselines}}
  \label{fig:SOTA offline-online RL baselines}
\end{figure}

\section{Conclusion}\label{sec_conclusion}

This work investigates the order-matching problem for coordinated ride-pooling and transit services under a dynamic and stochastic setting. We formulate an MDP model that encodes riders' drop-off location choices in agent action space and vehicle occupancy in agent states. We determine order-matching decisions by solving a bipartite matching problem periodically. The value for matching a rider and ride-pooling vehicle is determined based on both instant rewards and future gains from matching.  To learn the value function across spatial and temporal dimensions, we introduce an innovative RL framework RG-CQL, which combines offline training and online fine-tuning for enhancing training efficiency and effectiveness. The offline training employs CDDQN as the potential agent executor and extracts valuable insights from batches of past transitions concerning environmental dynamics. The online training stage fine-tuned the algorithms through real-world interactions. Noticing the optimism-pessimism gap, we devise a Guider, which is trained at the offline stage through supervised learning and enables CDDQN to explore unknown high-reward actions at the online fine-tuning stage. Simulation experiments based on Manhattan taxi trip data reveal that pooling-transit services notably boost order service rates and platform profits compared to other service modes, such as pure ride-pooling. The results also showcase the remarkable efficiency and effectiveness of the proposed RG-CQL framework. With an initial exploration rate set at 10\%, implementing the RG-CQL method reduces online training time by over 81\%, improves oveall performance by 4.3\%, and reduce overestimation by 5.6\%, compared to the online RL method DDQN with sufficient exploration. {Furthermore, thanks to the introduction of Guider, our RG-CQL also manages to mitigate the pessimism-optimism gap between offline RL and online fine-tuning in context of coordinating ride-pooling with public transit, over-performing SOTA offline to online RL baselines (CQL, Cal-q, and Hybrid-q) in  terms of both total accumulative rewards and training efficiency.}



Future research can explore the following avenues. First, while our current focus is on using ride-pooling to solve the transit first-mile problem, we have overlooked the last-mile issue. A valuable research direction would be to investigate the order-matching problem when ride-pooling is applied to address both the first and last-mile challenges of transit services.
Second, enhancing the Guider through improved transition predictions offers a promising research direction.
Third, more operational strategies in ride-hailing platforms could be incorporated into the RG-CQL framework, such as { considerations of multi-agent interdependencies}, vehicle repositioning issues, discriminatory pricing for passengers, and optimal wage setting for drivers.
  
\section*{Acknowledgments}  {This research was supported by Hong Kong Research Grants Council under project 16202922 and 26200420.}


\appendix 
\section{Notations} \label{Appendix_notations}
\nomenclature[S]{\(\mathcal{N}\)}{Set of pooling vehicles}
\nomenclature[S]{\(\mathcal{M}\)}{Set of unserved riders}
\nomenclature[S]{$\mathcal{T}$}{Whole planning horizon of the system}
\nomenclature[S]{$\mathcal{I}$}{Transit stations set}
\nomenclature[S]{$\mathcal{Z}$}{Set of city zones}
\nomenclature[S]{\(K_{t}\)}{Set of time steps afterwards t until the end of the planning horizon}
\nomenclature[S]{\(X\)}{Set of decision variables of ILP}
\nomenclature[S]{$\mathcal{D}$}{Set of past data transitions}
\nomenclature[S]{$A(s_{n,t})$}{Set of random actions offered by Guider under state \(s_{n,t}\)} 

\nomenclature[F]{\(R_{t}\)}{Summation of rewards of all N agents at time t}
\nomenclature[F]{\(r_{n,t}\)}{Reward received by vehicle n at time t}
\nomenclature[F]{\(P\)}{Transition Function of the environment}
\nomenclature[F]{\(Q_{\Pi}\)}{Q value of the overall platform under centralized policy}
\nomenclature[F]{\(\Pi\)}{Centralized policy that maps the state space to the action space}
\nomenclature[F]{\(\Pi^*\)}{Centralized optimal policy that maps the state space to the action space}
\nomenclature[F]{\(p\)}{Transition Function of individual agent}
\nomenclature[F]{\(Q_{n}\)}{Q value of agent n}
\nomenclature[F]{\(\pi\)}{Individual policy shared by all agent}
\nomenclature[F]{\(\pi^*\)}{Optimal individual policy shared by all agent}
\nomenclature[F]{\(L\)}{Loss function of DDQN}
\nomenclature[F]{\(L_c\)}{Loss function of CDDQN}
\nomenclature[F]{\(L_g\)}{Loss function of Guider model}

\nomenclature[P]{\(n\)}{Vehicle n}
\nomenclature[P]{\(m\)}{Matched rider m}
\nomenclature[P]{\(t\)}{Decision time t}
\nomenclature[P]{$i$}{Transit station i}
\nomenclature[P]{\(l_{n,t}\)}{Current location of vehicle n at time t}
\nomenclature[P]{\(v_{n,t}\)}{Number of vacant seats of vehicle n at time t}
\nomenclature[P]{\(p_{n,t}\)}{Information of passengers on board of vehicle n at time t}
\nomenclature[P]{\(o_{m}\)}{Origin information of rider m}
\nomenclature[P]{\(d_{m}\)}{Destination information of rider m}
\nomenclature[P]{\(e_{m}\)}{Service request time of rider m}
\nomenclature[P]{\(l_{m}\)}{Expected arrival time of rider m}
\nomenclature[P]{$z$}{City zone z}
\nomenclature[P]{$I_z$}{Transit stations within city zone z}
\nomenclature[P]{\(s_{n,t}\)}{State of vehicle n at time t}
\nomenclature[P]{\(c_{m}\)}{Seat capacity of a vehicle}
\nomenclature[P]{$i_k$}{The drop-off destination of kth passenger on board}
\nomenclature[P]{$t_k$}{The estimated remaining time in vehicle of kth passenger on board}
\nomenclature[P]{$\delta_k$}{The additional travel time of kth passenger on board as opposed to non-pooling services}
\nomenclature[P]{\(S_{t}\)}{Collections of state of all N agents at time t}
\nomenclature[P]{\(\beta_0\)}{Flag fare for a vehicle serving a new passenger}
\nomenclature[P]{\(\beta_1\)}{Cost per unit distance}
\nomenclature[P]{\(\beta_2\)}{Cost per unit waiting time for a rider}
\nomenclature[P]{\(\beta_3\)}{Cost per unit additional travel time for a rider within delay threshold}
\nomenclature[P]{\(\beta_4\)}{Cost per unit additional travel time for a rider beyond delay threshold}
\nomenclature[P]{$\kappa$}{Delay tolerance of the passenger}
\nomenclature[P]{$\Omega_{n,t}$}{Set of existing passengers on board of vehicle n at time t}
\nomenclature[P]{\(dis(o,d)\)}{O-D Euclidean distance of a a passenger order picked up by the vehicle}
\nomenclature[P]{\(\gamma\)}{Discount factor of the environment}
\nomenclature[P]{\(w_{n,m}\)}{Weight for matching rider m to vehicle n}
\nomenclature[P]{$\overline{Q}$}{Upper bound of Q value}
\nomenclature[P]{\(d_{n,m}\)}{Distance between vehicle n and order m}
\nomenclature[P]{\(R_{match}\)}{Maximum matching distance between vehicle and order}
\nomenclature[P]{\(\epsilon\)}{Exploration rate}
\nomenclature[P]{\(D\)}{Batch size of experience memories}
\nomenclature[P]{\(\tau\)}{Vehicle trajectory}
\nomenclature[P]{\(s\)}{Current state of vehicle trajectory}
\nomenclature[P]{\(a\)}{Action taken of vehicle trajectory}
\nomenclature[P]{\(r\)}{Reward received of vehicle trajectory}
\nomenclature[P]{\(s'\)}{Next state of vehicle trajectory}
\nomenclature[P]{\(C\)}{Hyper-parameter of conservative regularization term}
\nomenclature[P]{\(\alpha_c\)}{Learning rate of CDDQN}
\nomenclature[P]{\(\rho\)}{Soft update rate}
\nomenclature[P]{\(G\)}{Estimation of the Guider network}
\nomenclature[P]{\(\alpha_g\)}{Learning rate of Guider}
\nomenclature[P]{\(\widehat{r}\)}{Predefined reward threshold}
\nomenclature[P]{\(\beta\)}{Exploration decay rate}
\nomenclature[P]{\(\epsilon_T\)}{Predefined final exploration rate}
\nomenclature[P]{\(M\)}{Experience sample size}
\nomenclature[P]{\(p_{pool}\)}{Percentage of passengers who have a strong preference for pooling-only services}

\nomenclature[D]{\(A_{t}\)}{Collections of action of all N agents at time t}
\nomenclature[D]{\(a_{n,t}\)}{Action of vehicle n at time t}
\nomenclature[D]{\(x_{n,m}\)}{Matching decision for a specific vehicle n-order m pair}
\nomenclature[D]{\(\theta\)}{Neural network parameters of the training Q network}
\nomenclature[D]{\(\theta^-\)}{Neural network parameters of the target Q network}
\nomenclature[D]{\(\phi\)}{Neural network parameters of the Guider network}
\renewcommand{\nomname}{}
\printnomenclature

\bibliographystyle{unsrt}
\bibliography{sample-base}

\begin{thebibliography}{10}

\bibitem{schaller2017empty}
Bruce Schaller.
\newblock Empty seats, full streets: Fixing manhattan’s traffic problem.
\newblock {\em Schaller Consulting}, 1(3):1--27, 2017.

\bibitem{chen2018pricing}
Yiwei Chen and Hai Wang.
\newblock Pricing for a last-mile transportation system.
\newblock {\em Transportation Research Part B: Methodological}, 107:57--69, 2018.

\bibitem{wang2016approximating}
Hai Wang and Amedeo Odoni.
\newblock Approximating the performance of a “last mile” transportation system.
\newblock {\em Transportation Science}, 50(2):659--675, 2016.

\bibitem{wang2019routing}
Hai Wang.
\newblock Routing and scheduling for a last-mile transportation system.
\newblock {\em Transportation Science}, 53(1):131--147, 2019.

\bibitem{qin2022government}
Xiaoran Qin, Jintao Ke, and Hai Yang.
\newblock Government regulations for ride-sourcing services as substitute or complement to public transit.
\newblock {\em Available at SSRN 4129034}, 2022.

\bibitem{singh2021distributed}
Ashutosh Singh, Abubakr~O Al-Abbasi, and Vaneet Aggarwal.
\newblock A distributed model-free algorithm for multi-hop ride-sharing using deep reinforcement learning.
\newblock {\em IEEE Transactions on Intelligent Transportation Systems}, 23(7):8595--8605, 2021.

\bibitem{haliem2021adapool}
Marina Haliem, Vaneet Aggarwal, and Bharat Bhargava.
\newblock Adapool: A diurnal-adaptive fleet management framework using model-free deep reinforcement learning and change point detection.
\newblock {\em IEEE Transactions on Intelligent Transportation Systems}, 23(3):2471--2481, 2021.

\bibitem{wang2023optimization}
Dujuan Wang, Qi~Wang, Yunqiang Yin, and TCE Cheng.
\newblock Optimization of ride-sharing with passenger transfer via deep reinforcement learning.
\newblock {\em Transportation Research Part E: Logistics and Transportation Review}, 172:103080, 2023.

\bibitem{feng2022coordinating}
Siyuan Feng, Peibo Duan, Jintao Ke, and Hai Yang.
\newblock Coordinating ride-sourcing and public transport services with a reinforcement learning approach.
\newblock {\em Transportation Research Part C: Emerging Technologies}, 138:103611, 2022.

\bibitem{gao2024regulating}
Jing Gao and Sen Li.
\newblock Regulating for-hire autonomous vehicles for an equitable multimodal transportation network.
\newblock {\em Transportation Research Part B: Methodological}, 183:102925, 2024.

\bibitem{xu2024design}
Meng Xu, Yining Di, Zheng Zhu, Hai Yang, and Xiqun Chen.
\newblock Design and analysis of ride-sourcing services with auxiliary autonomous vehicles for transportation hubs in multi-modal transportation systems.
\newblock {\em Transportmetrica B: Transport Dynamics}, 12(1):2333869, 2024.

\bibitem{wang2024coordinative}
Xiaohan Wang, Xiqun~Michael Chen, Chi Xie, and Taesu Cheong.
\newblock Coordinative dispatching of shared and public transportation under passenger flow outburst.
\newblock {\em Transportation Research Part E: Logistics and Transportation Review}, 189:103655, 2024.

\bibitem{stiglic2018enhancing}
Mitja Stiglic, Niels Agatz, Martin Savelsbergh, and Mirko Gradisar.
\newblock Enhancing urban mobility: Integrating ride-sharing and public transit.
\newblock {\em Computers \& Operations Research}, 90:12--21, 2018.

\bibitem{ma2019dynamic}
Tai-Yu Ma, Saeid Rasulkhani, Joseph~YJ Chow, and Sylvain Klein.
\newblock A dynamic ridesharing dispatch and idle vehicle repositioning strategy with integrated transit transfers.
\newblock {\em Transportation Research Part E: Logistics and Transportation Review}, 128:417--442, 2019.

\bibitem{gu2024algorithms}
Qian-Ping Gu and Jiajian~Leo Liang.
\newblock Algorithms and computational study on a transportation system integrating public transit and ridesharing of personal vehicles.
\newblock {\em Computers \& Operations Research}, 164:106529, 2024.

\bibitem{sutton1998introduction}
Richard~S Sutton, Andrew~G Barto, et~al.
\newblock {\em Introduction to reinforcement learning}, volume 135.
\newblock MIT press Cambridge, 1998.

\bibitem{kumar2020conservative}
Aviral Kumar, Aurick Zhou, George Tucker, and Sergey Levine.
\newblock Conservative q-learning for offline reinforcement learning.
\newblock {\em Advances in Neural Information Processing Systems}, 33:1179--1191, 2020.

\bibitem{yu2022batch}
Xinlian Yu and Song Gao.
\newblock A batch reinforcement learning approach to vacant taxi routing.
\newblock {\em Transportation Research Part C: Emerging Technologies}, 139:103640, 2022.

\bibitem{tang2019deep}
Xiaocheng Tang, Zhiwei Qin, Fan Zhang, Zhaodong Wang, Zhe Xu, Yintai Ma, Hongtu Zhu, and Jieping Ye.
\newblock A deep value-network based approach for multi-driver order dispatching.
\newblock In {\em Proceedings of the 25th ACM SIGKDD International Conference on Knowledge Discovery \& Data Mining}, pages 1780--1790, 2019.

\bibitem{jiao2021real}
Yan Jiao, Xiaocheng Tang, Zhiwei~Tony Qin, Shuaiji Li, Fan Zhang, Hongtu Zhu, and Jieping Ye.
\newblock Real-world ride-hailing vehicle repositioning using deep reinforcement learning.
\newblock {\em Transportation Research Part C: Emerging Technologies}, 130:103289, 2021.

\bibitem{wei2023reinforcement}
Honghao Wei, Zixian Yang, Xin Liu, Zhiwei Qin, Xiaocheng Tang, and Lei Ying.
\newblock A reinforcement learning and prediction-based lookahead policy for vehicle repositioning in online ride-hailing systems.
\newblock {\em IEEE Transactions on Intelligent Transportation Systems}, 2023.

\bibitem{levine2020offline}
Sergey Levine, Aviral Kumar, George Tucker, and Justin Fu.
\newblock Offline reinforcement learning: Tutorial, review, and perspectives on open problems.
\newblock {\em arXiv preprint arXiv:2005.01643}, 2020.

\bibitem{nakamoto2024cal}
Mitsuhiko Nakamoto, Simon Zhai, Anikait Singh, Max Sobol~Mark, Yi~Ma, Chelsea Finn, Aviral Kumar, and Sergey Levine.
\newblock Cal-ql: Calibrated offline rl pre-training for efficient online fine-tuning.
\newblock {\em Advances in Neural Information Processing Systems}, 36, 2024.

\bibitem{ziegler2019fine}
Daniel~M Ziegler, Nisan Stiennon, Jeffrey Wu, Tom~B Brown, Alec Radford, Dario Amodei, Paul Christiano, and Geoffrey Irving.
\newblock Fine-tuning language models from human preferences.
\newblock {\em arXiv preprint arXiv:1909.08593}, 2019.

\bibitem{ouyang2022training}
Long Ouyang, Jeffrey Wu, Xu~Jiang, Diogo Almeida, Carroll Wainwright, Pamela Mishkin, Chong Zhang, Sandhini Agarwal, Katarina Slama, Alex Ray, et~al.
\newblock Training language models to follow instructions with human feedback.
\newblock {\em Advances in Neural Information Processing Systems}, 35:27730--27744, 2022.

\bibitem{rafailov2024direct}
Rafael Rafailov, Archit Sharma, Eric Mitchell, Christopher~D Manning, Stefano Ermon, and Chelsea Finn.
\newblock Direct preference optimization: Your language model is secretly a reward model.
\newblock {\em Advances in Neural Information Processing Systems}, 36, 2024.

\bibitem{wang2018stable}
Xing Wang, Niels Agatz, and Alan Erera.
\newblock Stable matching for dynamic ride-sharing systems.
\newblock {\em Transportation Science}, 52(4):850--867, 2018.

\bibitem{agatz2011dynamic}
Niels Agatz, Alan~L Erera, Martin~WP Savelsbergh, and Xing Wang.
\newblock Dynamic ride-sharing: A simulation study in metro atlanta.
\newblock {\em Procedia-Social and Behavioral Sciences}, 17:532--550, 2011.

\bibitem{hsieh2023improving}
Fu-Shiung Hsieh.
\newblock Improving acceptability of cost savings allocation in ridesharing systems based on analysis of proportional methods.
\newblock {\em Systems}, 11(4):187, 2023.

\bibitem{hsieh2024comparison}
Fu-Shiung Hsieh.
\newblock Comparison of a hybrid firefly--particle swarm optimization algorithm with six hybrid firefly--differential evolution algorithms and an effective cost-saving allocation method for ridesharing recommendation systems.
\newblock {\em Electronics}, 13(2):324, 2024.

\bibitem{van2016deep}
Hado Van~Hasselt, Arthur Guez, and David Silver.
\newblock Deep reinforcement learning with double q-learning.
\newblock In {\em Proceedings of the AAAI Conference on Artificial Intelligence}, volume~30, 2016.

\bibitem{Ma2013TshareAL}
Shuo Ma, Yu~Zheng, and Ouri Wolfson.
\newblock T-share: A large-scale dynamic taxi ridesharing service.
\newblock {\em 2013 IEEE 29th International Conference on Data Engineering (ICDE)}, pages 410--421, 2013.

\bibitem{alonso2017demand}
Javier Alonso-Mora, Samitha Samaranayake, Alex Wallar, Emilio Frazzoli, and Daniela Rus.
\newblock On-demand high-capacity ride-sharing via dynamic trip-vehicle assignment.
\newblock {\em Proceedings of the National Academy of Sciences}, 114(3):462--467, 2017.

\bibitem{simonetto2019real}
Andrea Simonetto, Julien Monteil, and Claudio Gambella.
\newblock Real-time city-scale ridesharing via linear assignment problems.
\newblock {\em Transportation Research Part C: Emerging Technologies}, 101:208--232, 2019.

\bibitem{ke2023supply}
Jintao Ke, Hai Yang, Hai Wang, and Yafeng Yin.
\newblock {\em Supply and demand management in ride-sourcing markets}.
\newblock Elsevier, 2023.

\bibitem{8206203}
Javier Alonso-Mora, Alex Wallar, and Daniela Rus.
\newblock Predictive routing for autonomous mobility-on-demand systems with ride-sharing.
\newblock In {\em 2017 IEEE/RSJ International Conference on Intelligent Robots and Systems (IROS)}, pages 3583--3590, 2017.

\bibitem{tsao2019model}
Matthew Tsao, Dejan Milojevic, Claudio Ruch, Mauro Salazar, Emilio Frazzoli, and Marco Pavone.
\newblock Model predictive control of ride-sharing autonomous mobility-on-demand systems.
\newblock In {\em 2019 International conference on robotics and automation (ICRA)}, pages 6665--6671. IEEE, 2019.

\bibitem{ali2023rebalancing}
Muhammad~Sajid Ali, Nagacharan~Teja Tangirala, Alois Knoll, and David Eckhoff.
\newblock Rebalancing autonomous electric vehicles for mobility-on-demand by data-driven model predictive control.
\newblock In {\em 2023 IEEE 26th International Conference on Intelligent Transportation Systems (ITSC)}, pages 215--221. IEEE, 2023.

\bibitem{shah2020neural}
Sanket Shah, Meghna Lowalekar, and Pradeep Varakantham.
\newblock Neural approximate dynamic programming for on-demand ride-pooling.
\newblock In {\em Proceedings of the AAAI Conference on Artificial Intelligence}, volume~34, pages 507--515, 2020.

\bibitem{yu2019integrated}
Xian Yu and Siqian Shen.
\newblock An integrated decomposition and approximate dynamic programming approach for on-demand ride pooling.
\newblock {\em IEEE Transactions on Intelligent Transportation Systems}, 21(9):3811--3820, 2019.

\bibitem{you2024approximate}
Fan You and Thomas Vossen.
\newblock An approximate dynamic programming approach to dynamic stochastic matching.
\newblock {\em INFORMS Journal on Computing}, 2024.

\bibitem{luo2023efficient}
Qi~Luo, Viswanath Nagarajan, Alexander Sundt, Yafeng Yin, John Vincent, and Mehrdad Shahabi.
\newblock Efficient algorithms for stochastic ride-pooling assignment with mixed fleets.
\newblock {\em Transportation Science}, 2023.

\bibitem{mnih2015human}
Volodymyr Mnih, Koray Kavukcuoglu, David Silver, Andrei~A Rusu, Joel Veness, Marc~G Bellemare, Alex Graves, Martin Riedmiller, Andreas~K Fidjeland, Georg Ostrovski, et~al.
\newblock Human-level control through deep reinforcement learning.
\newblock {\em Nature}, 518(7540):529--533, 2015.

\bibitem{vinyals2019grandmaster}
Oriol Vinyals, Igor Babuschkin, Wojciech~M Czarnecki, Micha{\"e}l Mathieu, Andrew Dudzik, Junyoung Chung, David~H Choi, Richard Powell, Timo Ewalds, Petko Georgiev, et~al.
\newblock Grandmaster level in starcraft ii using multi-agent reinforcement learning.
\newblock {\em Nature}, 575(7782):350--354, 2019.

\bibitem{berner2019dota}
Christopher Berner, Greg Brockman, Brooke Chan, Vicki Cheung, Przemys{\l}aw D{\k{e}}biak, Christy Dennison, David Farhi, Quirin Fischer, Shariq Hashme, Chris Hesse, et~al.
\newblock Dota 2 with large scale deep reinforcement learning.
\newblock {\em arXiv preprint arXiv:1912.06680}, 2019.

\bibitem{al2019deeppool}
Abubakr~O Al-Abbasi, Arnob Ghosh, and Vaneet Aggarwal.
\newblock Deeppool: Distributed model-free algorithm for ride-sharing using deep reinforcement learning.
\newblock {\em IEEE Transactions on Intelligent Transportation Systems}, 20(12):4714--4727, 2019.

\bibitem{zhu2020analysis}
Zheng Zhu, Xiaoran Qin, Jintao Ke, Zhengfei Zheng, and Hai Yang.
\newblock Analysis of multi-modal commute behavior with feeding and competing ridesplitting services.
\newblock {\em Transportation Research Part A: Policy and Practice}, 132:713--727, 2020.

\bibitem{liu2021mobility}
Yining Liu and Yanfeng Ouyang.
\newblock Mobility service design via joint optimization of transit networks and demand-responsive services.
\newblock {\em Transportation Research Part B: Methodological}, 151:22--41, 2021.

\bibitem{fan2024optimal}
Wenbo Fan, Weihua Gu, and Meng Xu.
\newblock Optimal design of ride-pooling as on-demand feeder services.
\newblock {\em Transportation Research Part B: Methodological}, 185:102964, 2024.

\bibitem{peng2019advantage}
Xue~Bin Peng, Aviral Kumar, Grace Zhang, and Sergey Levine.
\newblock Advantage-weighted regression: Simple and scalable off-policy reinforcement learning.
\newblock {\em arXiv preprint arXiv:1910.00177}, 2019.

\bibitem{fujimoto2019off}
Scott Fujimoto, David Meger, and Doina Precup.
\newblock Off-policy deep reinforcement learning without exploration.
\newblock In {\em International Conference on Machine Learning}, pages 2052--2062. PMLR, 2019.

\bibitem{fujimoto2021minimalist}
Scott Fujimoto and Shixiang~Shane Gu.
\newblock A minimalist approach to offline reinforcement learning.
\newblock {\em Advances in Neural Information Processing Systems}, 34:20132--20145, 2021.

\bibitem{kostrikov2021offline}
Ilya Kostrikov, Ashvin Nair, and Sergey Levine.
\newblock Offline reinforcement learning with implicit q-learning.
\newblock In {\em International Conference on Learning Representations}, 2022.

\bibitem{chebotar2023q}
Yevgen Chebotar, Quan Vuong, Karol Hausman, Fei Xia, Yao Lu, Alex Irpan, Aviral Kumar, Tianhe Yu, Alexander Herzog, Karl Pertsch, et~al.
\newblock Q-transformer: Scalable offline reinforcement learning via autoregressive q-functions.
\newblock In {\em Conference on Robot Learning}, pages 3909--3928. PMLR, 2023.

\bibitem{kumar2022offline}
Aviral Kumar, Rishabh Agarwal, Xinyang Geng, George Tucker, and Sergey Levine.
\newblock Offline q-learning on diverse multi-task data both scales and generalizes.
\newblock In {\em International Conference on Learning Representations}, 2023.

\bibitem{he2016deep}
Kaiming He, Xiangyu Zhang, Shaoqing Ren, and Jian Sun.
\newblock Deep residual learning for image recognition.
\newblock In {\em Proceedings of the IEEE conference on computer vision and pattern recognition}, pages 770--778, 2016.

\bibitem{vaswani2017attention}
Ashish Vaswani, Noam Shazeer, Niki Parmar, Jakob Uszkoreit, Llion Jones, Aidan~N Gomez, {\L}ukasz Kaiser, and Illia Polosukhin.
\newblock Attention is all you need.
\newblock {\em Advances in Neural Information Processing Systems}, 30, 2017.

\bibitem{song2022hybrid}
Yuda Song, Yifei Zhou, Ayush Sekhari, Drew Bagnell, Akshay Krishnamurthy, and Wen Sun.
\newblock Hybrid rl: Using both offline and online data can make rl efficient.
\newblock In {\em International Conference on Learning Representations}, 2023.

\bibitem{lee2022offline}
Seunghyun Lee, Younggyo Seo, Kimin Lee, Pieter Abbeel, and Jinwoo Shin.
\newblock Offline-to-online reinforcement learning via balanced replay and pessimistic q-ensemble.
\newblock In {\em Conference on Robot Learning}, pages 1702--1712. PMLR, 2022.

\bibitem{mark2022fine}
Max~Sobol Mark, Ali Ghadirzadeh, Xi~Chen, and Chelsea Finn.
\newblock Fine-tuning offline policies with optimistic action selection.
\newblock In {\em Deep Reinforcement Learning Workshop NeurIPS 2022}, 2022.

\bibitem{GoogleImages}
Google Search.
\newblock Google images, https://www.google.com.

\bibitem{boscoe2012nationwide}
Francis~P Boscoe, Kevin~A Henry, and Michael~S Zdeb.
\newblock A nationwide comparison of driving distance versus straight-line distance to hospitals.
\newblock {\em The Professional Geographer}, 64(2):188--196, 2012.

\bibitem{boyaci2021vehicle}
Burak Boyac{\i}, Thu~Huong Dang, and Adam~N Letchford.
\newblock Vehicle routing on road networks: How good is euclidean approximation?
\newblock {\em Computers \& Operations Research}, 129:105197, 2021.

\bibitem{tang2021value}
Xiaocheng Tang, Fan Zhang, Zhiwei Qin, Yansheng Wang, Dingyuan Shi, Bingchen Song, Yongxin Tong, Hongtu Zhu, and Jieping Ye.
\newblock Value function is all you need: A unified learning framework for ride hailing platforms.
\newblock In {\em Proceedings of the 27th ACM SIGKDD Conference on Knowledge Discovery \& Data Mining}, pages 3605--3615, 2021.

\bibitem{sadeghi2022reinforcement}
Soheil Sadeghi~Eshkevari, Xiaocheng Tang, Zhiwei Qin, Jinhan Mei, Cheng Zhang, Qianying Meng, and Jia Xu.
\newblock Reinforcement learning in the wild: Scalable rl dispatching algorithm deployed in ridehailing marketplace.
\newblock In {\em Proceedings of the 28th ACM SIGKDD Conference on Knowledge Discovery and Data Mining}, pages 3838--3848, 2022.

\bibitem{fujimoto2018addressing}
Scott Fujimoto, Herke Hoof, and David Meger.
\newblock Addressing function approximation error in actor-critic methods.
\newblock In {\em International Conference on Machine Learning}, pages 1587--1596. PMLR, 2018.

\bibitem{oda2018movi}
Takuma Oda and Carlee Joe-Wong.
\newblock Movi: A model-free approach to dynamic fleet management.
\newblock In {\em IEEE INFOCOM 2018-IEEE Conference on Computer Communications}, pages 2708--2716. IEEE, 2018.

\bibitem{haliem2021distributed}
Marina Haliem, Ganapathy Mani, Vaneet Aggarwal, and Bharat Bhargava.
\newblock A distributed model-free ride-sharing approach for joint matching, pricing, and dispatching using deep reinforcement learning.
\newblock {\em IEEE Transactions on Intelligent Transportation Systems}, 22(12):7931--7942, 2021.

\bibitem{NYCTaxiData2018}
NYC Taxi and Limousine Commission.
\newblock Nyc taxi and limousine commission-trip record data nyc, https://www1.nyc.gov/.

\bibitem{OpenStreetMap}
OpenStreetMap.
\newblock Openstreetmap, https://www.openstreetmap.org.

\bibitem{NYCTravelTimeMap}
Nate Parrott.
\newblock Nyc subway travel time map, https://subway.nateparrott.com.

\bibitem{MTASchedules}
Metropolitan~Transportation Authority.
\newblock Mta schedules, https://new.mta.info/schedules.

\bibitem{Scipy}
{Scipy}.
\newblock {Scipy library, https://scipy.org/}, 2024.

\bibitem{crouse2016implementing}
David~F Crouse.
\newblock On implementing 2d rectangular assignment algorithms.
\newblock {\em IEEE Transactions on Aerospace and Electronic Systems}, 52(4):1679--1696, 2016.

\bibitem{ProjectOSRM}
OSRM.
\newblock Project osrm, https://project-osrm.org.

\bibitem{OSRMAPI}
OSRM.
\newblock Project osrm api documentation, https://project-osrm.org/docs/v5.5.1/api.

\bibitem{Openrouteservice}
Openroute Service.
\newblock Project openroute service, https://openrouteservice.org/.

\bibitem{GoogleMaps}
Google.
\newblock Google maps, https://www.google.com/maps.

\bibitem{edirimanna2024integrating}
Danushka Edirimanna, Hins Hu, and Samitha Samaranayake.
\newblock Integrating on-demand ride-sharing with mass transit at-scale.
\newblock {\em arXiv preprint arXiv:2404.07691}, 2024.

\end{thebibliography}
\end{document}